\documentclass{elsarticle}
\usepackage{lineno,url,verbatim,graphicx,subfig,amsmath,import}
\graphicspath{ {./images/} }
\usepackage[margin=2.5cm]{geometry}%
\usepackage[T1]{fontenc}
\usepackage{ae}
\usepackage{aecompl}
\usepackage{graphicx}
\usepackage{enumitem}
\usepackage{tikz}
\usepackage{float}
\usepackage{multirow}

\usetikzlibrary{matrix,chains,positioning,decorations.pathreplacing,arrows}
\begin{document}

\begin{frontmatter}

\title{An Explainable Machine Learning Model for Early Detection of Parkinson's Disease using LIME on DaTscan Imagery}

\author{Pavan Rajkumar Magesh \qquad Richard Delwin Myloth \qquad Rijo Jackson Tom*
\\Dept. of Computer Science and Engineering
\\CMR Institute of Technology, Bengaluru, India
\\\{pavanraj.m14, richarddelwin07\}@gmail.com
\\ *Corresponding author
\\ E-mail address : rijojackson@gmail.com
\\ Phone : +91 95001 91494}

\begin{abstract}
Parkinson's disease (PD) is a degenerative and progressive neurological condition. Early diagnosis can improve treatment for patients and is performed through dopaminergic imaging techniques like the SPECT DaTscan. In this study, we propose a machine learning model that accurately classifies any given DaTscan as having Parkinson's disease or not, in addition to providing a plausible reason for the prediction. This is kind of reasoning is done through the use of visual indicators generated using Local Interpretable Model-Agnostic Explainer (LIME) methods. DaTscans were drawn from the Parkinson's Progression Markers Initiative database and trained on a CNN (VGG16) using transfer learning, yielding an accuracy of 95.2\%, a sensitivity of 97.5\%, and a specificity of 90.9\%. Keeping model interpretability of paramount importance, especially in the healthcare field, this study utilises LIME explanations to distinguish PD from non-PD, using visual superpixels on the DaTscans. It could be concluded that the proposed system, in union with its measured interpretability and accuracy may effectively aid medical workers in the early diagnosis of Parkinson's Disease.
\end{abstract}


\end{frontmatter}

\section{Introduction}

Parkinson's disease (PD) is a brain and nervous system dysfunction which is neurodegenerative in nature. This means that the malady results in, or is characterized by the degeneration of the nervous system, especially the neurons in the brain. Parkinson's disease exists as one of the most common \emph{neurodegenerative} diseases, exceeded only by that of Alzheimer's. It predominately affects dopamine-producing \emph{dopaminergic} neurons in a particular region of the brain (Figure 1) called the  \emph{substantia nigra} \cite{cite1}. In Parkinson's disease, a patient loses the ability to retain these dopamine-producing neurons which causes a loss of control over any voluntary actions. This disease may lead to motor and non-motor symptoms such as tremors, slowed movement, sleep disorders, posture imbalance, depression and other subtle symptoms \cite{cite2}. There exists a variety of medical scans such as Magnetic Resonance Imaging (MRI), Functional Magnetic Resonance Imaging (fMRI), Positron Emission Tomography (PET), etc. but the Single-photon Emission Computed Tomography (SPECT) functional imaging technique is most widely used in European clinics for the premature diagnosis of Parkinson's disease \cite{cite3}. The SPECT image technique utilises \textsuperscript{123}I-FP-CIT also known as\textsuperscript{123}I-Ioflupane. This is a ligand that binds to the \emph{dopamine transporters} (Hence, also called SPECT DaTscan) in the striatum region of the brain, namely \emph{putamen} and \emph{caudate}, very efficiently and with high affinity \cite{cite4}. PD patients are marked with significantly smaller putamen and caudate regions due to the lack of dopaminergic neurons as can be seen in Figure 2. \par

\begin{figure}[!ht]
\center
\includegraphics[scale = 0.9]{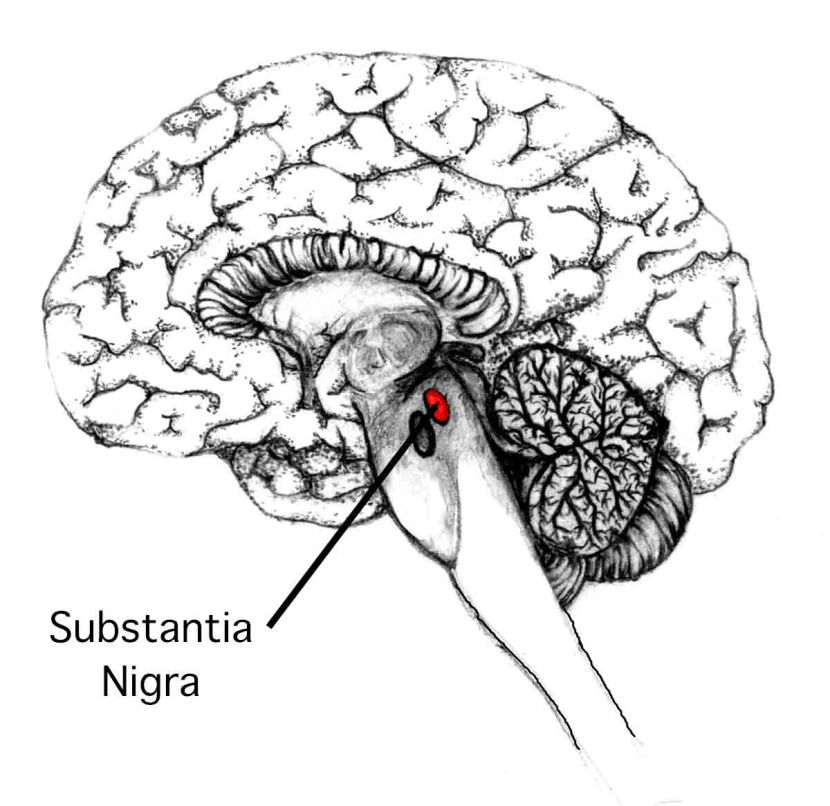}
\caption{Substania Nigra region of the brain}
\end{figure}

\begin{figure}[ht]
  \centering
  \subfloat[Healthy Control]{\includegraphics[scale = 0.37]{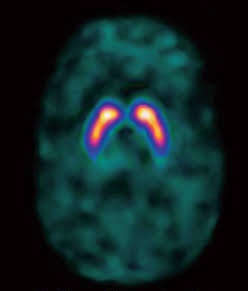}\label{fig:f1}}
\hspace{1 cm}
  \subfloat[Parkinson's Disease]{\includegraphics[scale = 0.37]{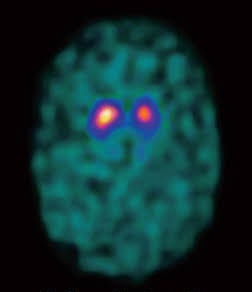}\label{fig:f2}}
  \caption{SPECT DaTscan with putamen and caudate regions marked by high contrast}
\end{figure}

The healthcare industry is a major and critical branch of the general service industry, where a bulk of the analysis of diagnostic data is performed by medical experts. The exposition of medical images is,  therefore, quite limited to specific experts having a profound knowledge of the subject and also due to the intricacy or variation of parameters amongst the data being handled. A neuro-image based diagnosis for Parkinson's may be appropriate considering that a symptomatic based treatment may come off as late and not be a time-conscious solution. Also, SPECT scan images are interpreted manually in clinics where the diagnosis result may be subject to the risk of human error. A previous clinical study found that the validity of diagnosis for PD, performed by movement disorder experts, was found initially to be 79.6\% and then rose to 83.9\% after follow-up checks which used DaTscans instead \cite{cite5}. \par 

Deep learning has widely been used for diagnosis  of various diseases and conditions, often with results exceeding standard benchmarks \cite{cite6}. Through the use of deep learning we can efficiently and accurately classify patients as to whether they have PD or not by detecting patterns in their SPECT scans, mainly around the putamen and caudate regions as they are relatively smaller as compared to non-PD specimens. Our work aims to provide an interpretable solution (using LIME - Local Interpretable Model-Agnostic Explanations) in addition to the binary classification result (PD or non-PD) of the developed black-box neural network so that medical experts may understand as to why the machine thinks this way, providing crucial insights for the decision making process. An overview of the experiment can be understood from Figure 3. \par
\bigskip
The main contributions of this paper are as follows:
\begin{itemize}[noitemsep]
  \item Development of an accurate deep learning model for the early diagnosis of Parkinson's Disease using SPECT DaTscans.
  \item Convey a comprehensive performance analysis of the VGG16 CNN model used for this medical imaging task.
  \item Provide an interpretable solution using LIME for the above classification problem.
  \item Aid medical practitioners in early diagnosis through the use of visual markings generated by the model on the predictions.
\end{itemize}

\bigskip

The remaining parts of the paper are assembled as follows - Section 2 discusses the Related Work performed in the field. The proposed system approach for the early detection of PD is explained in Section 3 and discusses the Dataset, Image Preprocessing, Dataset Splitting, Neural Network Architecture, Transfer Learning, and Results as subsections. Section 4 elaborates on the Explainability of the Proposed Model using LIME and discusses the Need for Interpretability, the LIME model, and the Interpretations of DaTscans as subsections. Section 5 finally discusses the Conclusions for this study.

\begin{figure}[!ht]
\includegraphics[width=\textwidth, height=15cm]{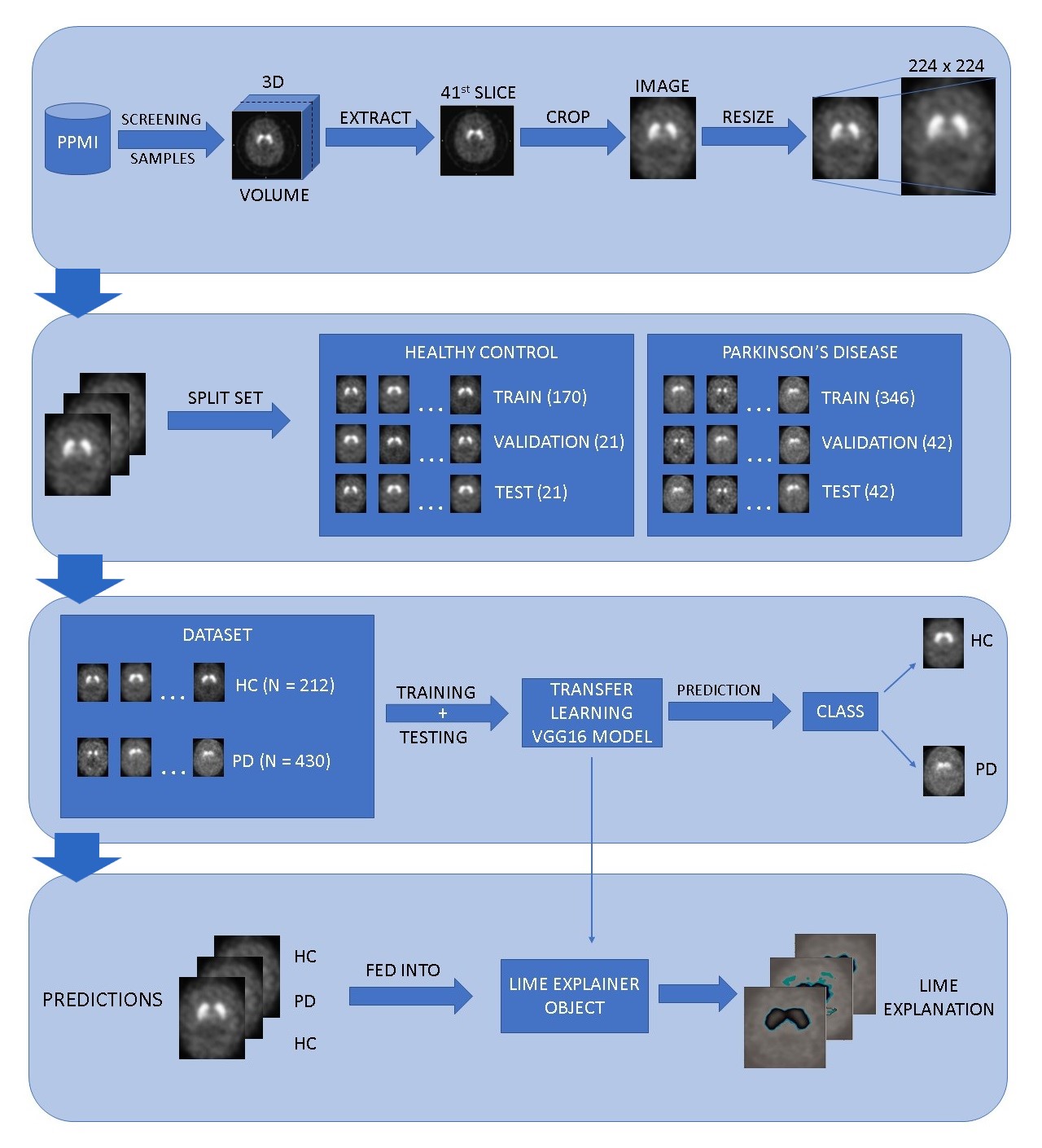}
\caption{Experiment Overview (PD = Parkinson's Disease, HC = Healthy Control, PPMI = Parkinson's Progression Marker's Initiative}
\end{figure}

\section{Related Work}

SPECT DaTscans are popularly utilised for the premature diagnosis of PD and have even been warranted by the Food and Drug Administration (FDA) in the United States. One of the earliest works to attempt to classify DaTscans as PD or non-PD was done by Towey et al. \cite{LS1}, where  Naive-Bayes was used with Principal Component Analysis (PCA) for the decomposition of voxels in the striatum region of the brain. Following this study Support Vector Machines (SVM) were utilised as the primary classifier mechanism with voxels as features (image voxel intensities). Such studies were conducted by Oliveira et al. \cite{LS2}. High accuracy classification was obtained by Prashanth et al. \cite{LS3} who used shape-surface based fitting and striatal binding ratio features along with SVMs. Apart from SPECT DaTscans, 3D MRI images were used by Cigdem et al. to classify PD using SVMs by comparing the morphological differences in the grey and white matter regions of the brain \cite{cigdem2018effects}. \par 

More recently deep learning based methods are being used in various fields of medical imaging as studied by Sheng et al. \cite{shen2017deep}. The use of Artificial Neural Networks (ANN) has been used to detect complex patterns in data and outperform classical statistical methods. Martinez-Murzia et al. \cite{LS4} and Rumman et al. \cite{rumman2018early} proposed the use of Convolutional Neural Networks (CNN) to detect patterns in DaTscan images associated with PD. Often 3D brain scans contain large amounts of details which can result in complex CNN architectures. Ortiz et al. \cite{ortiz2019parkinson} proposed the utilisation of iso-surfaces as a method to condense this consignment of data, simultaneously keeping the apposite details needed for classification. Limitations in compute capability prompted researchers like Quan et al. \cite{quan2019datscan} and Sivaranjini et al.  \cite{sivaranjini2019deep} to use transfer learning methods where weights and classification capabilities are transferred from existing popular CNN architectures \cite{simonyan2014very, krizhevsky2012imagenet, szegedy2016rethinking} to the model being developed for faster learning. \par

The question of explainability in healthcare has been long unanswered as most studies only attempt to achieve the highest accuracy metrics at their tasks, however, progress has been made to make these systems more interpretable and has been extensively studied by Ahmad et al. Furthermore \cite{ahmad2018interpretable}. Petsiuk et al., Lundberg et al. and Ribeiro et al. have proposed different frameworks \cite{petsiuk2018rise, lundberg2017unified, ribeiro2016should} for the interpretability of image classification problems that can be applied to medical images as well. Interpretable models for classification of other neurodegenerative diseases, such as Alzheimer's have been developed by Das et al. \cite{das2019interpretable} but none exist for Parkinson's Disease. This study attempts to close that gap by developing an explainable model for the same. \par

\section{Early Parkinson's Disease Detection CNN Model}

\subsection{\textbf{Dataset}}

The particulars utilised in this experiment were obtained from the \emph{Parkinson's Progression Markers Initiative (PPMI)} database \cite{cite7}. The PPMI is a surveying clinical study, whose inception resulted in the establishment of biomarker-defined cohorts used to identify genetic, clinical, imaging, and bio-specific progression markers. The study is funded by The Michael J. Fox Foundation for Parkinson's Research and is taking place in Europe, the United States, Australia, and Israel. \par

The dataset comprises of 642 DaTscan SPECT images divided into two classes namely PD (N = 430) and non-PD (N = 212). The data used was only from the initial screening of unique patients and no follow up scans of the same patient were used. This was done in accordance with the study's aim at \emph{early detection}, and also to maintain the uniqueness of the dataset. Another reason was to prevent any over-fitting while training the model, possibly caused due to any similarity between scans from the same patient. Scans without evidence for dopaminergic deficit (SWEDD) patients were also not included to maintain the integrity of the dataset. The demographics of the collected patient data is described in Table 1.

\begin{table}[ht]
\caption{Patient Demographics}
\centering
\begin{tabular}{ c c c }
\hline
Category & Healthy Control & Parkinson's Disease \\ [0.5ex] 
\hline
Number of patients & 212 & 430 \\
Sex (Male) & 128 &  278 \\
Sex (Female) & 84 & 152 \\
Age (Minimum) & 31 & 33 \\
Age (Maximum) & 84 & 85 \\
\hline
\end{tabular}
\label{table:demographics}
\end{table}

\subsection{\textbf{Image Preprocessing}}

The raw SPECT DaTscan images taken at PPMI affiliated medical clinics had undergone some preprocessing before they were added to the online database. Firstly, they went through attenuation correction procedures. This was achieved using phantoms procured from the same time the subject was imaged. Furthermore, they had been reconstructed and spatially normalized to eliminate any differences in shape or size  against several unique subjects. This alignment was done in accordance with the \emph{Montreal Neurological Institute (MNI)} accepted,  standard coordinate system for medical imaging. \par

Each $n^{th}$ SPECT DaTscan was finally presented as a 3D volume space, $(x_i^n,y_i^n,z_i^n)$ in \emph{DICOM}and \emph{NIFTI} format where \emph{i} represents the $i^{th}$ pixel on the \emph{x} and \emph{y} axes respectively. The $z$ axis represents the number of slices of the volume. Each $n^{th}$ volume can be represented as three sets of pixels on the $x,\ y,$ and $z$ axes. 

\bigskip

\begin{equation*}
 X^{(n)} = \{x_1^{(n)},x_2^{(n)},.... ,x_{91}^{(n)}\} 
\end{equation*}
\begin{equation*}
 Y^{(n)} = \{y_1^{(n)},y_2^{(n)},.... ,y_{109}^{(n)}\} 
\end{equation*}
\begin{equation*}
 Z^{(n)} = \{z_1^{(n)},z_2^{(n)},.... ,z_{91}^{(n)}\} 
\end{equation*}

This indicates that each volume had dimensions of 91 x 109 x 91, representing 91 slices with each slice being 109 x 91 pixels. \par

After visual analysis of the slices, keeping the putamen and caudate regions of the brain as the regions of interest (ROI), as well as referring to previous studies \cite{rumman2018early,quan2019datscan}, it was decided to use slice 41 i.e. $(x_i^n,y_i^n,z_{41}^n)$ for development as it depicted the ROI with the highest prominence. Hence the 41st slice of the DICOM image was extracted from all collected subject's data and converted to \emph{jpeg} format. Due to the varying sizes of male and female brain scans all images underwent \emph{cropping} to eliminate the black corners present for smaller brains, a characteristic observed mainly for female subjects. This process brought uniformity in size to all the scan images through the detection of the major contours and edges in the DaTscan. Due to the small dataset size, certain augmentations were applied to the training data to prevent over-fitting. These include height and width shifts, flips across the horizontal axis and brightness and intensity variations. The images were finally resized into 224 x 224 to be compatible with the \emph{VGG16} neural network architecture which we would be using.

\subsection{\textbf{Dataset Splitting}}

The dataset consisting of 642 images was divided into training, validation, and test sets in an 80:10:10 ratio respectively with each category being further divided into Healthy control and Parkinson's Disease (PD). The number of images in each set is summarised in Table 2.

\begin{table}[ht]
\caption{Dataset Split}
\centering
\begin{tabular}{ c | c c }
\hline
Category & Healthy Control & Parkinson's Disease \\ [0.5ex] 
\hline
Training & 170 & 346\\
Validation & 21 & 42 \\
Test & 21 & 42 \\
\hline
Total & 212 & 430 \\
\hline
\end{tabular}
\label{table:datasplit}
\end{table}

\subsection{\textbf{Neural Network Architecture}}

\emph{Deep learning} is furnishing inspiring solutions for medical image analysis issues and is seen as a leading method for ensuing applications in the field \cite{zhang2019radiological}. Deep Learning models utilise several layers of neural nodes to process its input. Each neuron collects a set of \emph{x-values} (vector) as input and quantifies the anticipated \emph{y-values}. Vector \emph{X} holds the worth of the features in one of the \emph{m} examples from the training set. Each of the units has their own assortment of parameters, usually referred to as \emph{w} (weights) and \emph{b} (bias) which undergoes changes during the learning or computation process. In every iteration, the neuron quantifies a weighted average of the values of the vector \emph{X}, which is based on its present weight vector \emph{W} and then adds bias. Finally, the outcome of this estimation is passed through a non-linear activation function $f_{act}$ as shown in Figure 4. \par

\bigskip

These deep learning models, in tasks like binary classification, often have performance exceeding even those of humans. In this study, we utilise  \emph{Convolutional Neural Networks (CNN)} that takes 2-dimensional or 3-dimensional shaped (i.e. structured) values as input. A CNN is a typical example of an Artificial Neural Network (ANN) positioned on conserving dimensional relationships among data. The inputs to a CNN are organized in a grid-like composition and further passed through layers that conserve these correspondences, each layer function working on a miniature zone of the previous layer. A CNN usually possesses several layers of activations and convolutions, distributed amongst pooling layers, and is trained by employing algorithms such as backpropagation and gradient descent. CNNs are generally structured such that in the end, they possess fully connected layers. Such layers are responsible for quantifying the final classes. All these layers constitute the basic building blocks of a CNN and can be visualised in Figure 5. Due to the systemic attributes of images, namely the configural features among bordering voxels or pixels, CNNs have achieved substantial popularity in medical image analysis \cite{shen2017deep}. \par

\begin{figure}[!ht]
\center
\includegraphics[scale = 0.47]{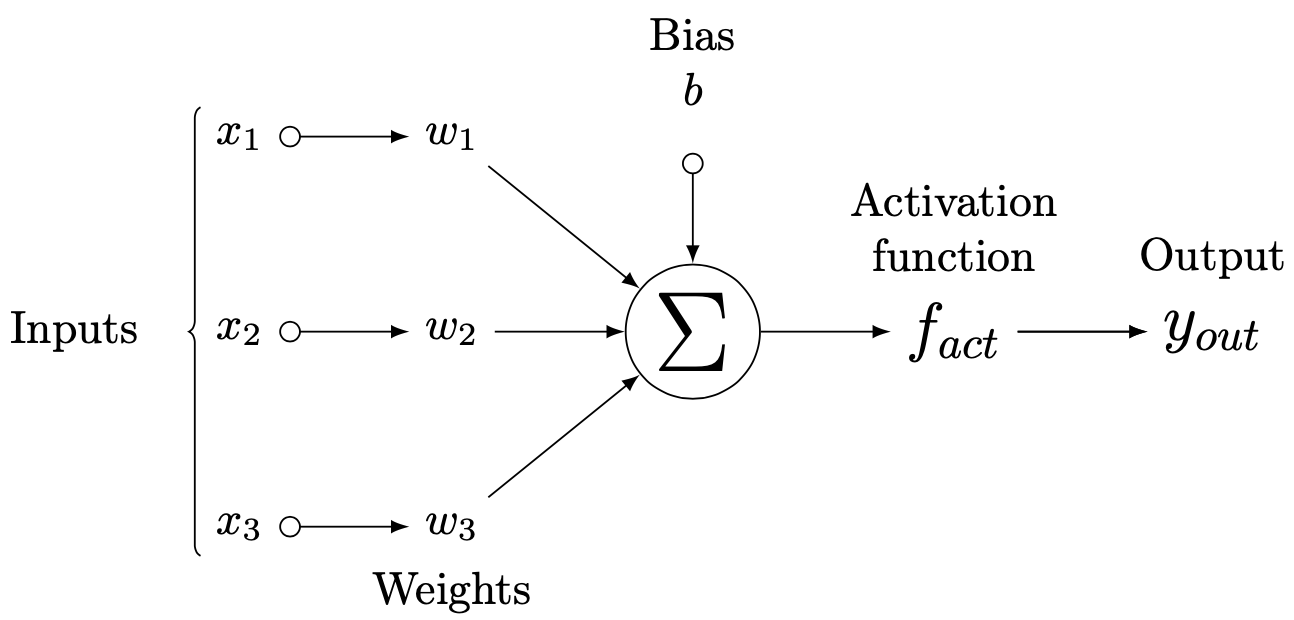}
\caption{Neural Network pipeline over each iteration}
\end{figure}

\begin{figure}[!ht]
\includegraphics[trim=0 0 0 250, clip, width=\textwidth]{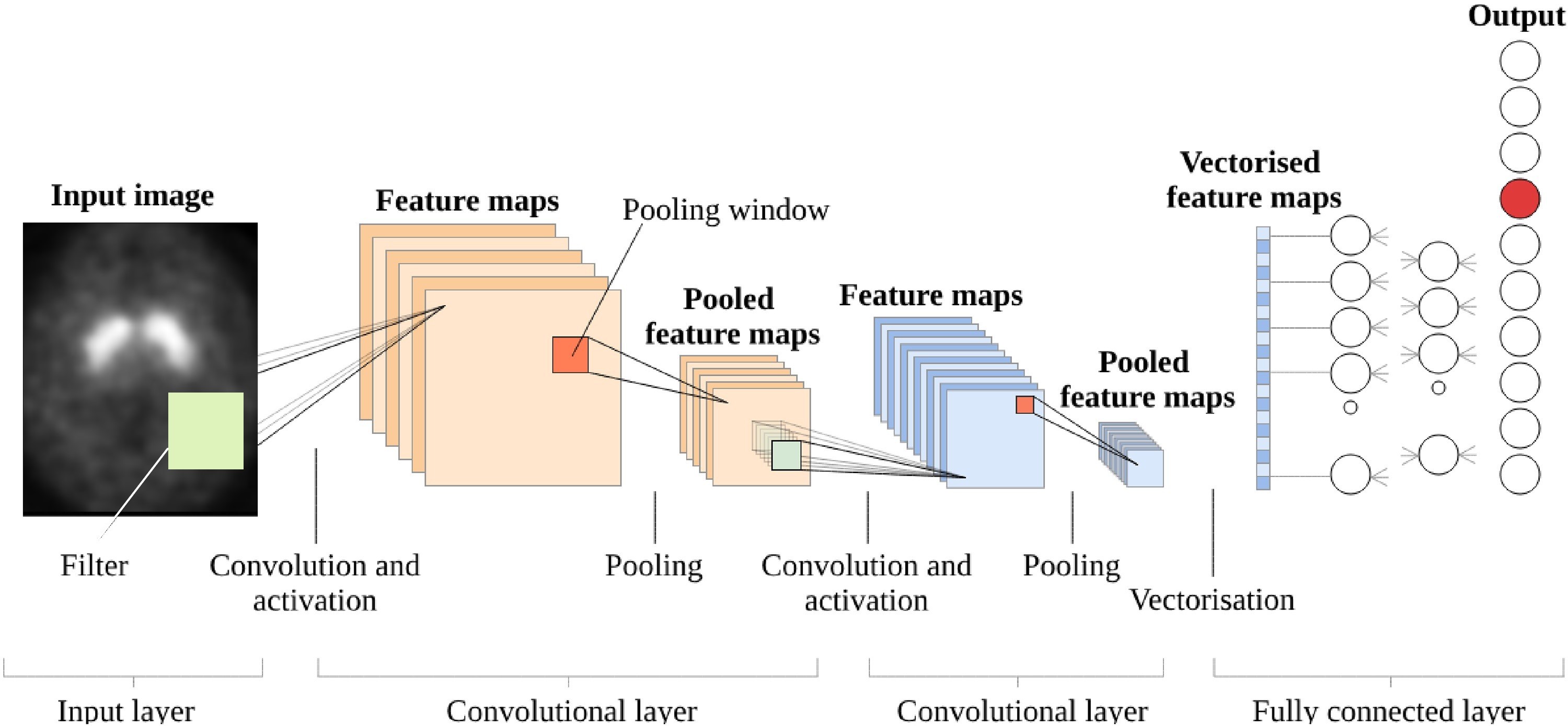}
\caption{Building Blocks of a typical CNN}
\end{figure}

The CNN used in this study is VGG16 \cite{simonyan2014very} which won the 2014 version of the \emph{ImageNet Large Scale Visual Recognition Challenge (ILSVRC)}. The model achieves a respectable 92.7\% test accuracy on ImageNet, which is a gigantic dataset of more than 1.2 million images attributed to 1000 classes. This was implemented using Keras, which is one of the leading high-level neural network APIs running on a Tensorflow backend. The model's layers consist of convolutional, max-pooling, activation and fully connected layers as shown in Figure 6. \par

\begin{figure}[ht]
\includegraphics[width=\textwidth]{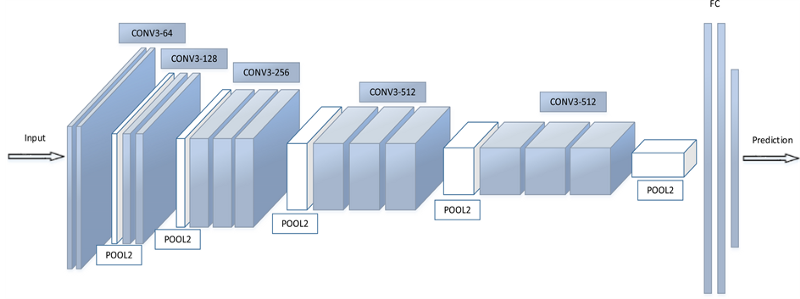}
\caption{VGG16 Architecture}
\end{figure}

The image is fed through a stack of convolutional layers. The filters were utilised with a minute responsive field of dimensions 3x3 units. This allows it to capture the smallest size conception in 4-way perpendicular directions. In a part of the organisation, the architecture also employs 1x1 convolutional filters. This can be perceived as a rectilinear transmutation of the input channels, and is trailed by non-rectilinear transformations. Spatial or dimensional pooling is executed by 5 max-pooling layers, that trail the convolutional layers. Thereafter, three fully connected layers trail a stack of convolutional layers that possesses varied depths. The initial two fully connected layers possess 4096 channels each while the last one conducts 1000-way ILSVRC classification and therefore has 1000 channels (specific only to the ImageNet classification task). The ultimate layer is the soft-max layer utilized for determining probabilities amongst several classes and uses the softmax function as shown in Equation 1.

\begin{equation}
P(y=j|z^i)=\phi_{softmax}(z^i)= \frac{e^{z^i}}{\sum_{j=0}^{k}e^{z_k^i}}
\end{equation}

where we define the net input \emph{z} as

\begin{equation}
z = w_0x_0+w_1x_1+....+w_mx_m = \displaystyle\sum_{l=0}^{m}w_lx_l = w^Tx
\end{equation}

\emph{w} and \emph{x} are the weight and feature vectors of a single training example, and where the bias unit is denoted as $w_0$. The softmax function quantifies the expectation that the training example $x^i$ is a member of the class \emph{j} on the basis of the weight and cumulative input $z^i$, and hence computes the probability \emph{p} for each class label in \textit{j = 1,2,...,k}. \par 

This base model was partially modified to  accommodate the needs of our study and the details are described in the next section.

\begin{table}[ht]
\caption{VGG16 Layers Table. Convolution layer is shortened to 'Conv.' its description includes: number of channels, kernel size; padding 'p'; and stride 'st'. Pooling layer is shortened to 'Pool'. Fully connected layer is shortened to 'FC'. Dropout layer is shortened to 'Drop'.}
\center
\includegraphics[scale = 0.5]{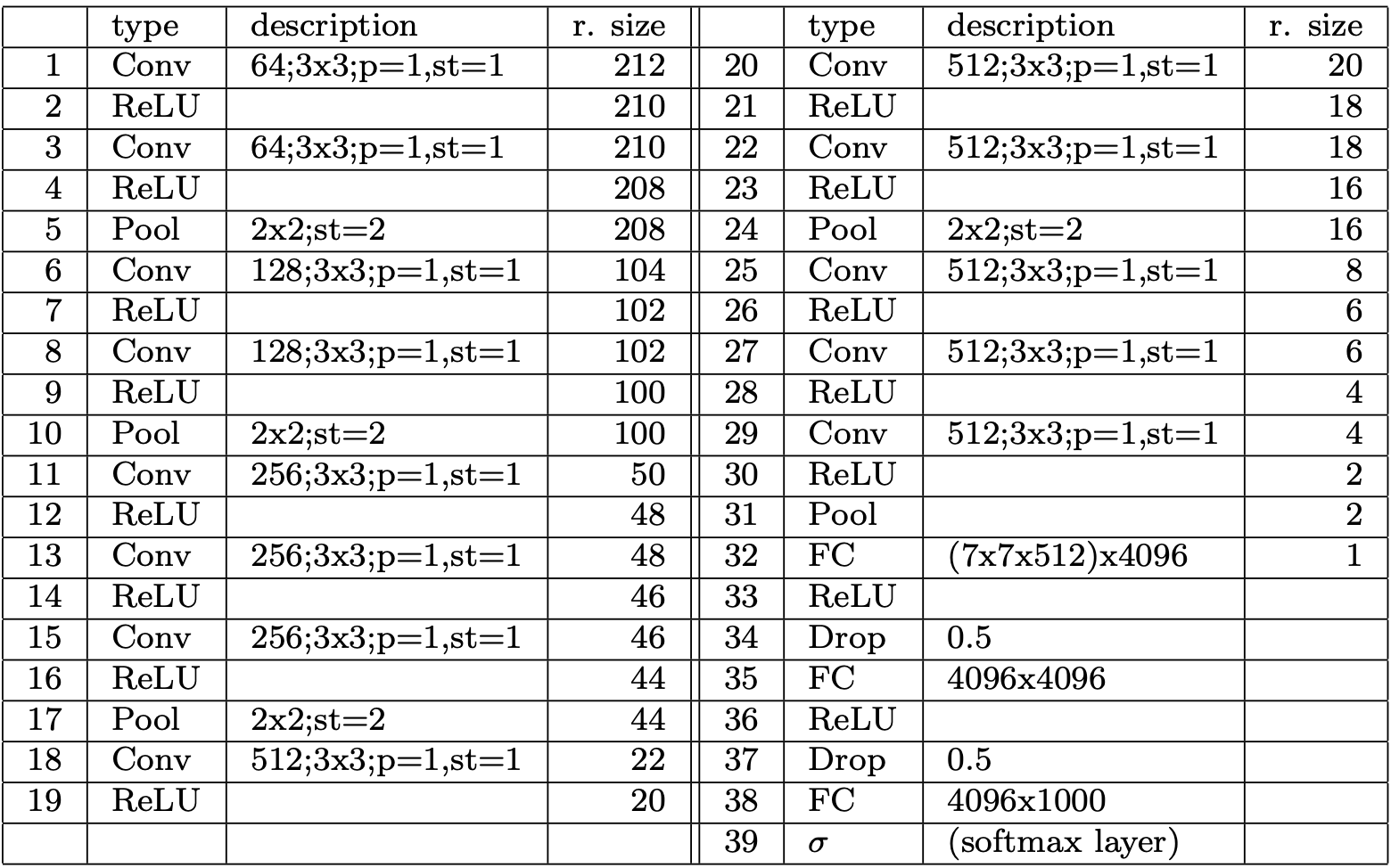}
\end{table}

\subsection{\textbf{Transfer Learning}}

In practice, it is unconventional to train a complete convolutional neural net from ground zero, especially with random values, since it is often infrequent to possess a dataset of substantial size or even the necessary computational resources to process it. To overcome this barrier, it is quite customary to pre-train a CNN on a sizeable dataset such as ImageNet, which as mentioned earlier, contains over a million images with 1000 classes \cite{deng2009imagenet}. After this initial large scale training, the CNN can be used either as an initialization model or a fixed feature extractor. This method is known as \emph{Transfer Learning} \cite{torrey2010transfer}. \par

A set of more formal definitions for understanding Transfer Learning was given by Pan et al. \cite{pan2009survey} as seen below: \par

\textbf{Definition 1} \textit{(Domain)} "A domain \textit{D = \{X,P(X)\}} is defined by two components: A feature space \emph{X} and a marginal probability distribution \textit{P(X)} where \textit{$X = {x_1,x_2,...,x_n}$}" \cite{torrey2010transfer}.
\par

\textbf{Definition 2} \textit{(Task)} "Given a specific domain \emph{D}, a task \textit{\{$Y,f(\cdot)$\}} consists of two parts: A label space \emph{Y} and a predictive function \textit{$f(\cdot)$}, which is not observed but can be learned from training data \textit{$\{(x_i,y_i) | i \in \{1,2,3......,N\}$, where $x_i \in X$ and $y_i \in Y\}$}" \cite{torrey2010transfer}. 

\par

\textbf{Definition 3} \textit{(Transfer Learning)} "Given a source domain \textit{$D_S$} and learning task \textit{$T_S$}, a target domain \textit{$D_T$} and learning task \textit{$T_T$}, \emph{transfer learning} aims to help improve the learning of the target predictive function \textit{$f_T(\cdot)$} in \textit{$D_T$} using the knowledge in \textit{$D_S$} and \textit{$T_S$}, where \textit{$D_S \neq D_T$} or \textit{$T_S \neq T_T$}" \cite{torrey2010transfer}. \par

Deep neural network models are prevalently layered systems that assimilate various features at different layers. Such layers are then ultimately connected to an end layer which is commonly fully connected, to retrieve the concluding output. Such layered systems permit us to make use of a prematurely trained network such as VGG16, short of its ultimate layer, as a fixed feature selector applicable to other recognition tasks. \par

In our study, however, we utilise a slightly more intricate methodology. Here, we not only replace the ultimate layer for task classification, but also particularly retrain a handful of the foregoing layers. As mentioned earlier, the inceptive layers have been observed to capture collective or non-specific features, while the later ones emphasise extensively on the particular task at hand. Utilising this discernment, we may \emph{freeze} (fix weights) particular layers while re-training, or \emph{fine-tune} the remaining to accommodate our requirements. The two last CNN layers of the stock VGG16 model were not frozen which allowed their weights to be trained specifically to the task at hand. In addition, two dropout layers and a single dense layer using a \emph{sigmoid} activation function were added to the end. The sigmoid activation function, also called the logistic function transforms the input to the function into a value between 0.0 and 1.0. This is especially helpful when we have to predict the probability as an output. \par
\begin{equation}
Sigmoid\  Function\  \phi(x) = \frac{1}{1 + e^{-x}}
\end{equation}

The simplicity in finding the derivative of the sigmoid function also helps in preparing the non-linear model of classification.
\begin{equation}
\frac{d}{dx}  \phi(x) = \phi(x)\cdot\phi(1-\phi(x))
\end{equation}

As a general summary, we make use of the knowledge or weights in terms of the comprehensive architecture of the neural net and hence utilise its states as  inception points for our retraining steps. This, as a result, helps us achieve superior execution through an improved rate of convergence resulting in smaller training times while requiring lesser memory for computation. \par
 
\subsection{\textbf{Results}}

The model was trained using an image sequence \textit{$X_t$}, where \textit{M = 516} (PD = 346, non-PD = 170) and validated using an image sequence \textit{$X_v$}, where \textit{N = 64} (PD = 42, non-PD = 21).

\begin{equation*}
 X_t = \{x_t^{(1)},x_t^{(2)},.... ,x_t^{(M)}\} 
\end{equation*}
\begin{equation*}
 X_v = \{x_v^{(1)},x_v^{(2)},.... ,x_v^{(N)}\} 
\end{equation*}

These sets used the corresponding class label sequences \textit{$Y_t$} and \textit{$Y_v$} respectively,

\begin{equation*}
 Y_t = \{y_t^{(1)},y_t^{(2)},.... ,y_t^{(M)}\} 
\end{equation*}
\begin{equation*}
 Y_v = \{y_v^{(1)},y_v^{(2)},.... ,y_v^{(N)}\} 
\end{equation*}

to effectively fit the distribution \textit{p(y)} by curtailing the cross-entropy using the loss function:

\begin{equation}
LE = - \frac{1}{N} \displaystyle\sum_{i=1}^{N} log(p(y_i))
\end{equation}

where \textit{N} represents the number of instances and \textit{$y_i$} depicts the classes being either a positive class (\textit{$y_1$}) or a negative class (\textit{$y_0$}). Mathematically speaking:

\begin{equation}
y_i = 1 \implies log(p(y_i))
\end{equation}
\begin{equation}
y_i = 0 \implies log(1-p(y_i))
\end{equation}

This yields the formula for the binary cross-entropy loss function as:

\begin{equation}
LE = - \frac{1}{N} \displaystyle\sum_{i=1}^{N} y_i \cdot log(p(y_i)) + (1-y_i) \cdot log(1-p(y_i))
\end{equation}

The training images underwent augmentations on the fly through the \emph{ImageDataGenerator} class with a training batch size of 32 and a validation batch size of 16. The model was trained over 300 epochs with a training step size of 32. The step size was decided using the general rule of thumb where the number of units in the dataset is divided by the batch size and the result obtained is multiplied by a positive integer greater than one, usually to account for augmentations. The validation step size was declared as 4 and estimated in a similar fashion. The optimizer in use was the \emph{Adam optimizer} of the \emph{Keras} optimizers library and the learning rate was initialized at 10$^{-3}$. The exponential decay rate, specifically for the first moment (beta 1) was set at 0.9 while that of the second-moment (beta 2) was set at 0.999 which must be near 1.0 for problems characterised by a sparse gradient, as in the case of computer vision. The total time taken for training on a cloud-based Tensor Processing Unit (TPU) device took 5460 seconds or roughly 1.5 hours. \par

\begin{table}[!ht]
\caption{Hyperparameters}
\centering
\begin{tabular}{ c c c }
\hline
Hyperparameter & Parameter Type & Value \\ [0.5ex] 
\hline
Epochs & - & 300\\
\multirow{2}{*}{Batch Size} & Training & 32\\
 & Validation & 16\\
 \multirow{2}{*}{Step Size} & Training & 32\\
 & Validation & 4\\
Learning Rate & - & $10^{-3}$\\
$\beta_1$ & First Moment & 0.9\\
$\beta_2$ & Second Moment & 0.999\\
Total Time & Seconds & 5460\\
\hline
\end{tabular}
\end{table}

\begin{equation}
Specificity = \frac{No.\ of\ True\ Negatives}{No.\ of\ True\ Negatives + No.\ of\ False\ Positives}
\end{equation}

\begin{equation}
Sensitivity = \frac{No.\ of\ True\ Positives}{No.\ of\ True\ Positives + No.\ of\ False\ Negatives}
\end{equation}

\begin{equation}
Precision = \frac{No.\ of\ True\ Positives}{No.\ of\ True\ Positives + No.\ of\ False\ Positives}
\end{equation}

The predictions on 64 test images (PD = 42, non-PD = 21) resulted in an \emph{accuracy} of 92.0\%, a \emph{specificity} of 81.8\%, a \emph{sensitivity} of 97.5\% and a \emph{precision} of 90.9\% estimated using the above formulae. A \emph{Cohen's Kappa} score of 0.81 and  \emph{F1} score of 0.94 were also obtained. The misclassified samples, which include 4 false positives and 1 false negative, can be seen in Figure 7. Important performance metrics are summarised in Table 5.

\begin{figure}[!ht]
  \subfloat{
	\begin{minipage}[c][1\width]{
	   0.145\textwidth}
	   \centering
	   \includegraphics[width=1\textwidth, height=2.4cm]{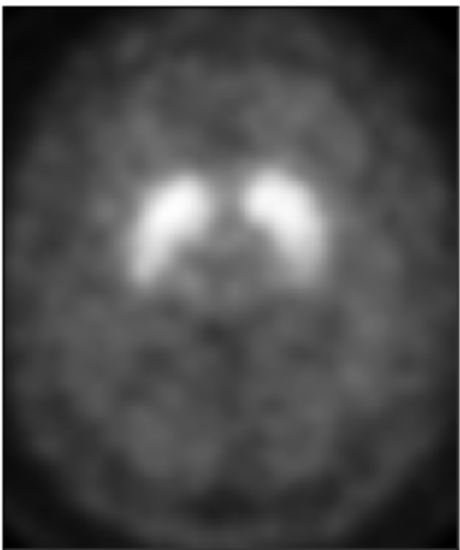}
	   \caption*{(a) False Positive}
	\end{minipage}}
 \hfill	
  \subfloat{
	\begin{minipage}[c][1\width]{
	   0.145\textwidth}
	   \centering
	   \includegraphics[width=1\textwidth, height=2.4cm]{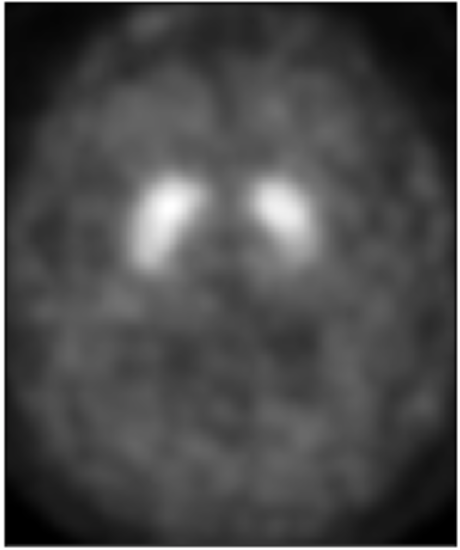}
	   \caption*{(b) False Positive}
	\end{minipage}}
\hfill	
  \subfloat{
	\begin{minipage}[c][1\width]{
	   0.145\textwidth}
	   \centering
	   \includegraphics[width=1\textwidth, height=2.4cm]{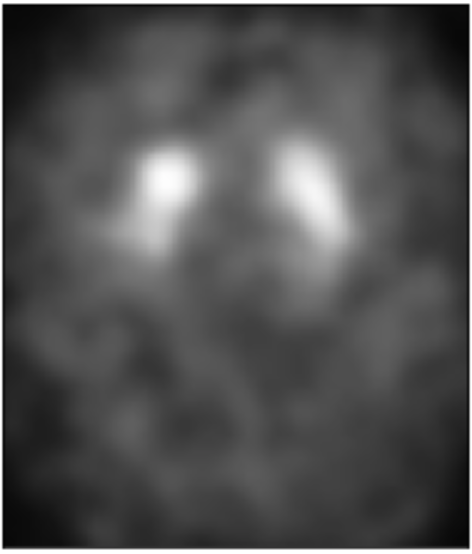}
	   \caption*{(c) False Positive}
	\end{minipage}}
\hfill	
  \subfloat{
	\begin{minipage}[c][\width]{
	   0.145\textwidth}
	   \centering
	   \includegraphics[width=\textwidth, height=2.4cm]{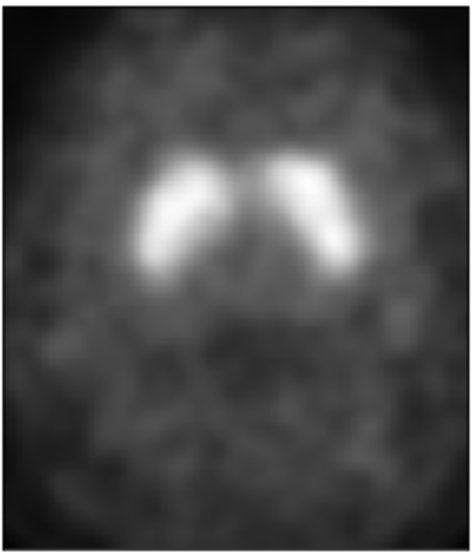}
	   \caption*{(d) False Positive}
	\end{minipage}}
\hfill	
  \subfloat{
	\begin{minipage}[c][\width]{
	   0.145\textwidth}
	   \centering
	   \includegraphics[width=1\textwidth, height=2.4cm]{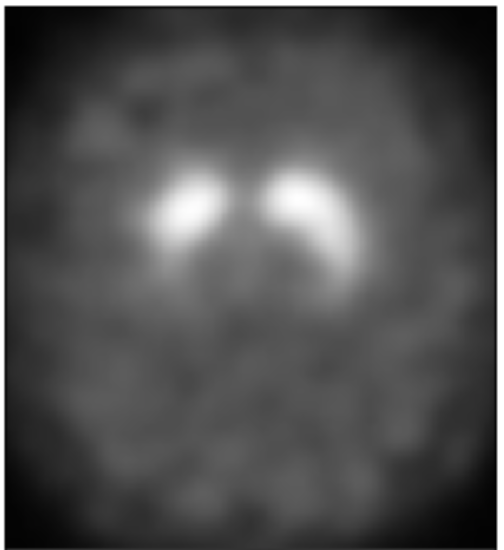}
	   \caption*{(e) False Negative}
	\end{minipage}}
	\bigskip
\caption{Misclassifications}
\end{figure}	

\begin{table}[!ht]
\caption{Performance Results}
\centering
\begin{tabular}{ c c | c c }
\hline
Category & Result & Metric & Result \\ [0.5ex] 
\hline
True Positives & 40 & Accuracy & 92.0\%\\
True Negatives & 18 & Specificity & 81.8\%\\
False Positives & 4 & Sensitivity & 97.5\%\\
False Negatives & 1 & Precision & 90.9\%\\
\hline
\end{tabular}
\label{table:results}
\end{table}

The progression of loss and accuracy over the number of epochs for the training and validation sets can be visualised in Figure 8(a) and Figure 8(b) respectively, and the confusion matrices for the model on the validation and test sets are shown in Figure 9(a) and Figure 9(b) respectively. \par

\begin{figure}[!ht]
  \centering
  \subfloat[Model Loss]{\includegraphics[scale = 0.33]{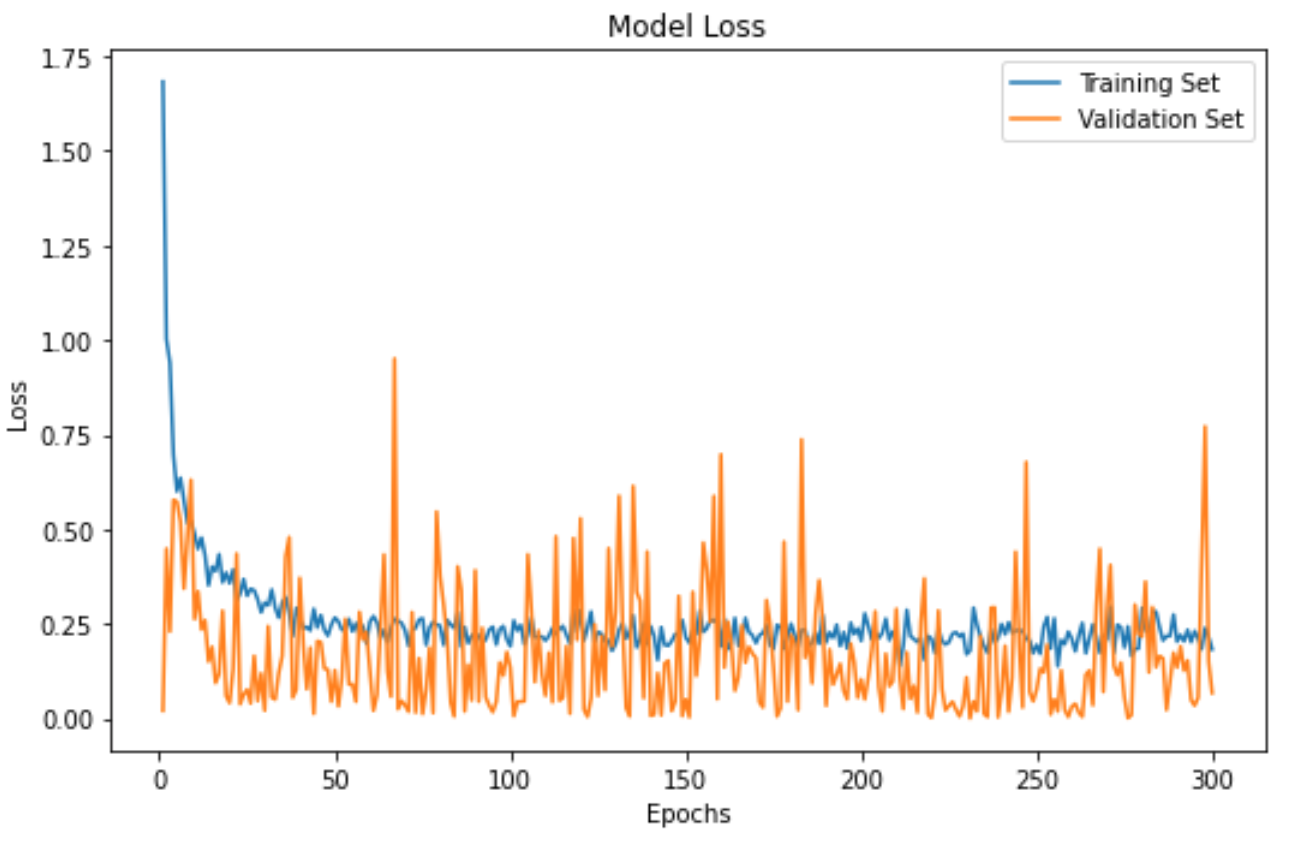}}
\hspace{1 cm}
  \subfloat[Model Accuracy]{\includegraphics[scale = 0.33]{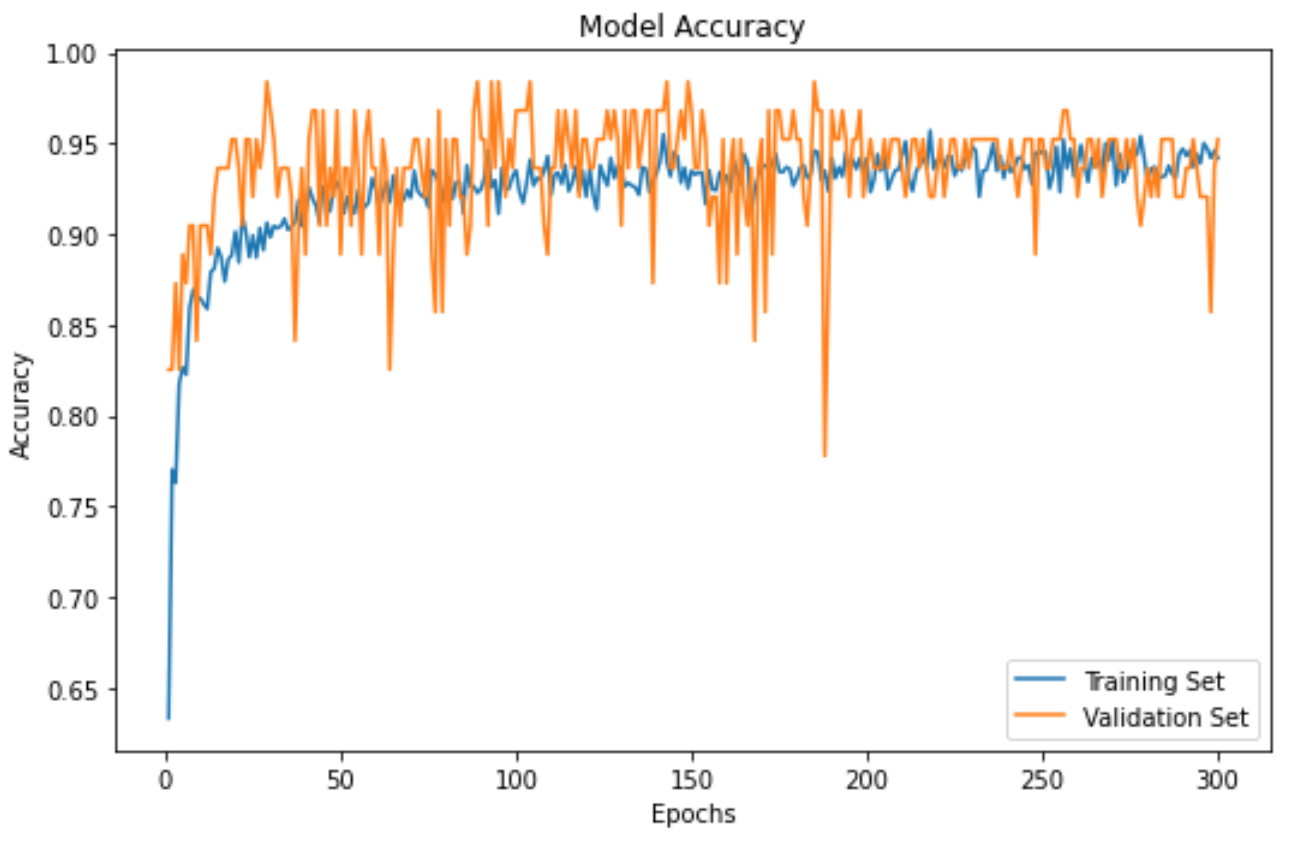}}
  \caption{Loss and Accuracy Progression}
\end{figure}

\begin{figure}[!ht]
  \centering
  \subfloat[Validation Set]{\includegraphics[scale = 0.37]{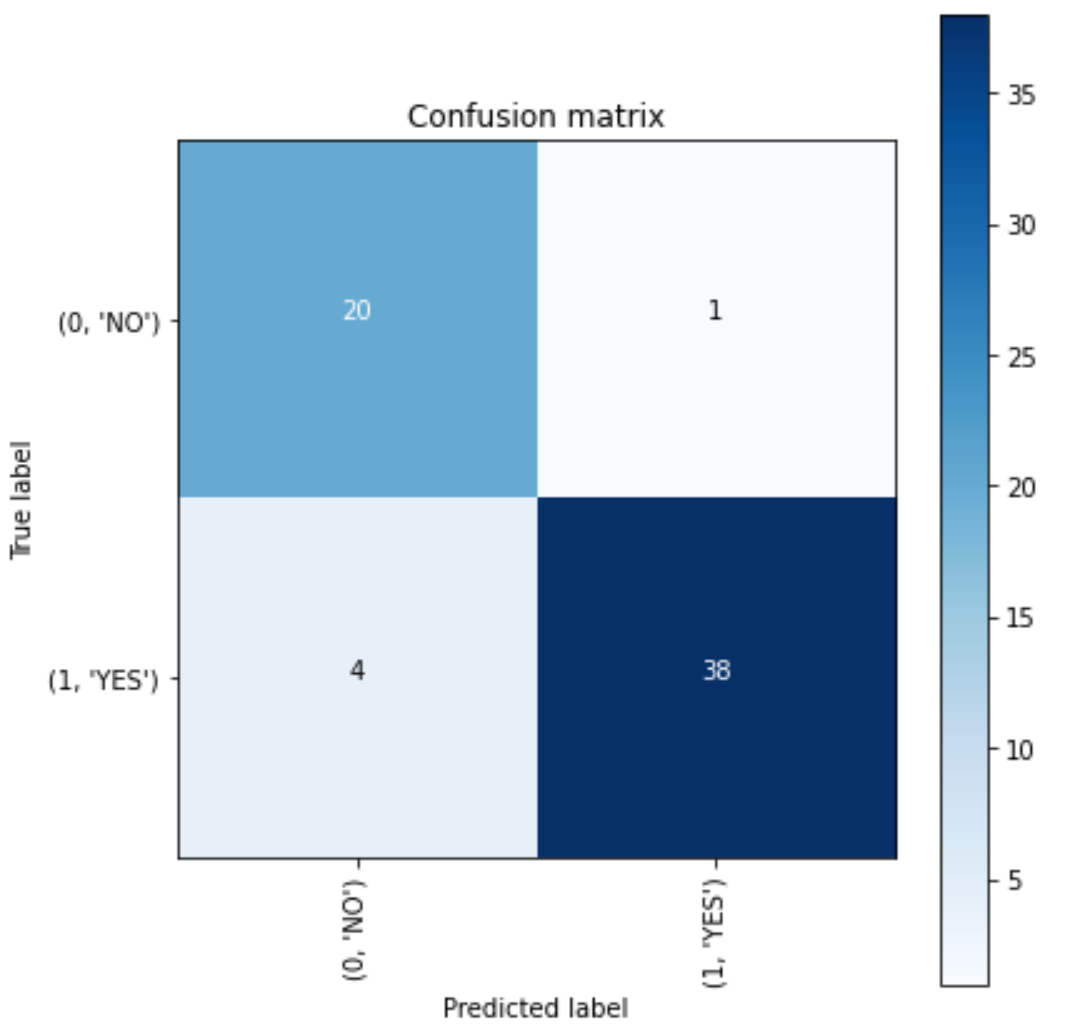}}
\hspace{2 cm}
  \subfloat[Test Set]{\includegraphics[scale = 0.37]{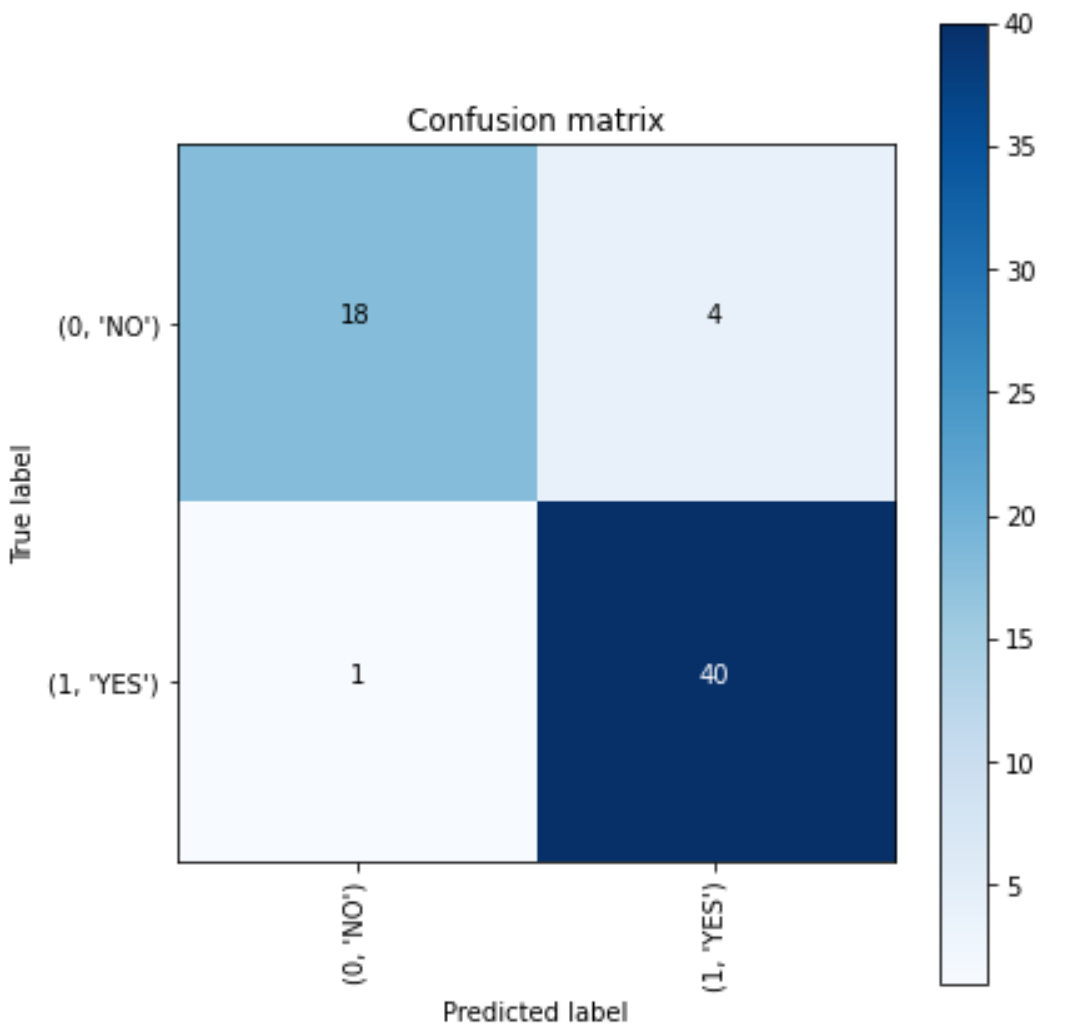}}
  \caption{Confusion Matrices}
\end{figure}

In this study the classification between PD and non-PD was obtained by normalizing the predicted probabilities by a parameter referred to as \emph{threshold} which was set at a value of 0.5, thus the values below the threshold of 0.5 are delegated to class 0 or non-PD and values above or equal to 0.5 are delegated to class 1 or PD. However, the default threshold may not be the ideal interpretation of the probabilities that have been predicted. Thus, the ROC (Receiver Operating Characteristic) curve is plotted to address these concerns as seen in Figure 10(b). The False Positive Rate, also abbreviated as FPR, is plotted on the horizontal axis while the True Positive Rate, also abbreviated as TPR, is plotted on the vertical axis. The dotted orange diagonal line on the plot which spans from the bottom-left to top-right of the figure indicates the curve for a no-skill classifier. The area under ROC (AUC) was found to be 0.89.

Analysing the curve gives insight to understanding the trade-off between the TPR and FPR for differing thresholds. The ROC Table (Table 6) shows the \emph{Geometric mean (G-mean)} values, where the higher values positively correlate to the better threshold values. The G-mean value is given by:

\begin{equation}
G_{mean} = \sqrt{Sensitivity * Specificity}
\end{equation}

Where sensitivity is the True Positive Rate (TPR) and specificity is the True Negative Rate (TNR). \par

The threshold value of 0.8335 was found to provide the best performance (indicated by the black dot present on the ROC curve). The \emph{Youden's J statistic} also known as the \emph{Youden index}, is a statistic that captures the performance of a binary class diagnostic test and further verifies the threshold value. The Youden statistic is given by:

\begin{equation}
Y = Sensitivity + Specificity - 1
\end{equation}

Coupled with the G-mean and Youden Index is the positive Likelihood Ratio (LR+) which is used in medical testing to interpret diagnostic tests and indicates how probable a patient possesses a disease. The positive LR depicts how much to multiply the probability of possessing a disease, given a positive test result. This ratio is given by:

\begin{equation}
LR+ = \frac{True\ Positive\ Rate\ (TPR)}{False\ Positive\ Rate\ (FPR)}
\end{equation}

A Similar well-known statistic used to determine the optimal threshold is the Precision-Recall (PR) curve as seen in Figure 10(a), which focuses on the performance of the model on the positive class, essentially indicating its' ability at predicting the positive class accurately. A no-skill model is represented by a horizontal line. The \emph{F-measure} is calculated to further strike the best balance between precision and recall. It is given by:

\begin{equation}
F_{measure} = \frac{2 * Precision * Recall}{Precision + Recall}
\end{equation}

The PR Table (Table 7) shows the F-scores for corresponding precision and recall values. Similar to G-mean, higher values of F-measure is a direct indication of the best threshold value corresponding to it, which in this case was observed as 0.8334. Thus, both ROC and PR analysis indicate that the optimal threshold value is 0.833. \par

\begin{figure}[!ht]
  \centering
  \subfloat[PR Curve]{\includegraphics[scale = 0.3]{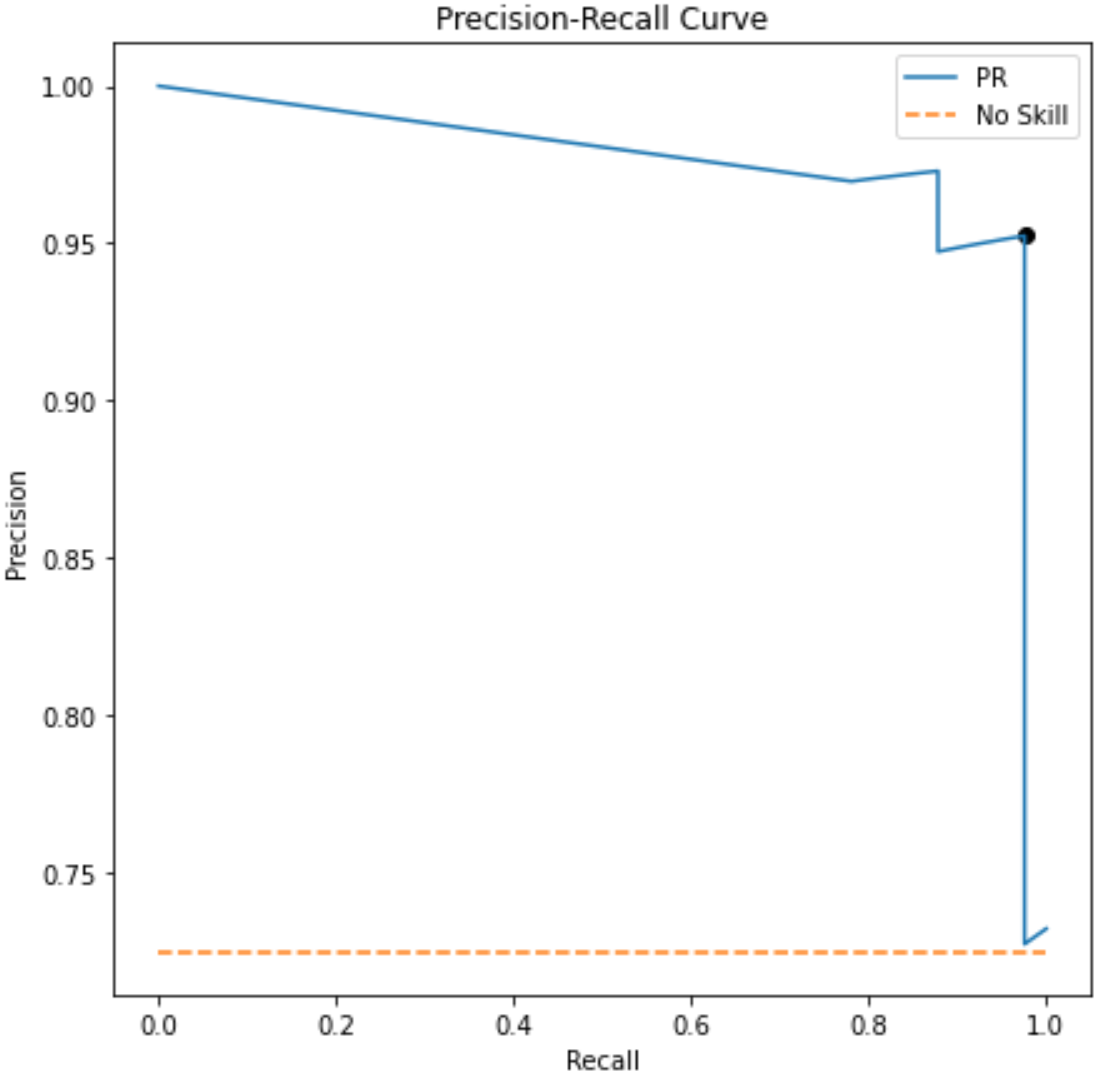}}
\hspace{1 cm}
  \subfloat[ROC Curve]{\includegraphics[scale = 0.3]{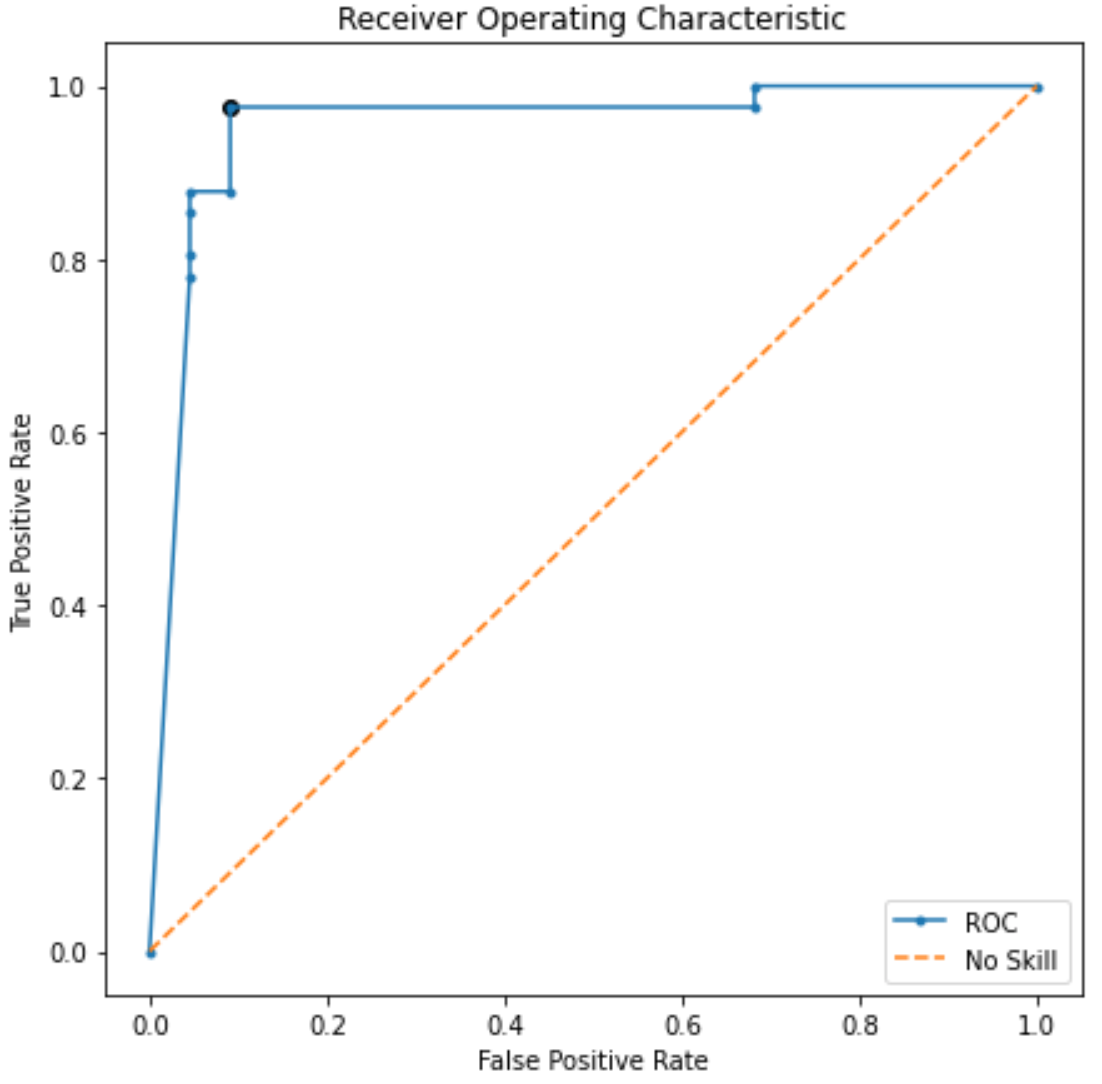}}
  \caption{Metric Curves}
\end{figure}

\begin{table}[!ht]
\caption{ROC Table}
\centering
\begin{tabular}{ c  c  c  c  c  c  c  c }
\hline
No. & Threshold & TPR (Sensitivity) & FPR (Fall-out) & Specificity & LR+ & Youden Index & G-mean \\ [0.5ex] 
\hline
1	& 2	& 0 &	0	&1	&-	&0&0\\
2&	1&	0.7805	&0.04545&	0.9545	&17.17&	0.735&		0.8631\\
3	&1	&0.8049&	0.04545	&0.9545	&17.71	&0.7594	&	0.8765\\
4	&1&	0.8537&	0.04545&	0.9545	&18.78	&0.8082&0.9027\\
5	&1	&0.878	&0.04545	&0.9545&	19.32&	0.8326	&0.9155\\
6	&1	&0.878&	0.09091&	0.9091&	9.659&	0.7871	&0.8934\\
\textbf{7}	&\textbf{0.8335}	&\textbf{0.9756}	&\textbf{0.09091}	&\textbf{0.9091}	&\textbf{10.73}	&\textbf{0.8847}	&	\textbf{0.9418}\\
8	&8.815e-08	&0.9756	&0.6818	&0.3182	&1.431	&0.2938	&	0.5572\\
9	&3.193e-08&	1	&0.6818	&0.3182	&1.467	&0.3182		&0.5641\\
10	&2.918e-10	&1&	1	&0	&1	&0	&0\\
\hline
\end{tabular}
\end{table}

\begin{table}[!ht]
\caption{PR Table}
\centering
\begin{tabular}{ c  c  c  c  c }
\hline
No. & Threshold & Precision & Recall & F-measure \\ [0.5ex] 
\hline
1 & 3.193e-08 & 0.7321 & 1 & 0.8453\\
2 & 8.814e-08 & 0.7272 & 0.9756 & 0.8333\\
3 & 1.865e-07 & 0.7407 & 	0.9756	 & 0.8421\\
4 & 4.164e-07 & 0.7547	 & 0.9756	 & 0.8510\\
5 & 5.929e-07 & 0.7692 & 	0.9756 & 	0.8602\\
6 & 1.611e-06 & 0.7843 & 	0.9756	 & 0.8695\\
7 & 1.130e-05 & 0.8000	 & 0.9756	 & 0.8791\\
8 & 1.851e-05 & 0.8163	 & 0.9756	 & 0.8888\\
9 & 0.0003 & 0.8333	 & 0.9756	 & 0.8988\\
10 & 0.0019 & 0.8510	 & 0.9756 & 	0.9090\\
11 & 0.0051	& 0.8695	 & 0.9756 & 	0.9195\\
12 & 0.0617 & 0.8888	 & 0.9756 & 	0.9302\\
13 & 0.6687 & 0.9090	 & 0.9756 & 	0.9411\\
14 & 0.8135 & 0.9302 & 	0.9756 & 	0.9523\\
\textbf{15} & \textbf{0.8334} & \textbf{0.9523}	 & \textbf{0.9756}	 & \textbf{0.9638}\\
16 & 0.9985	& 0.9512	 & 0.9512	 & 0.9512\\
17 & 0.9992 & 0.9500 & 	0.9268 & 	0.9382\\
18 & 0.9999	 & 0.9487	 & 0.9024	 & 0.9250\\
19 & 0.9999 & 0.9473	 & 0.8780	 & 0.9113\\
20 & 0.9999	 & 0.9729 & 	0.8780	 & 0.9230\\
21 & 1 & 0.9722 & 0.8536  & 0.9090\\
22 & 1  & 0.9705 & 0.8048	 & 0.8800\\
23 & 1 & 0.9696 & 0.7804	 & 0.8648\\
\hline
\end{tabular}
\end{table}

\begin{figure}[!ht]
  \centering
  \subfloat[ROC Table Representation]{\includegraphics[scale = 0.5]{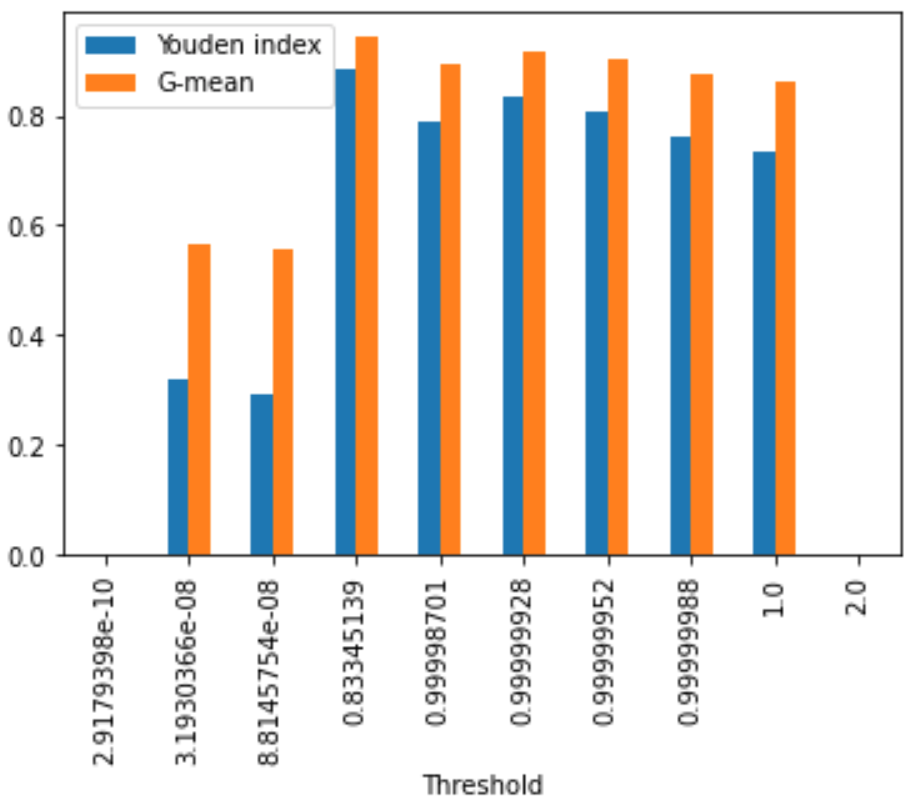}}
\hspace{2 cm}
  \subfloat[PR Table Representation]{\includegraphics[scale = 0.5]{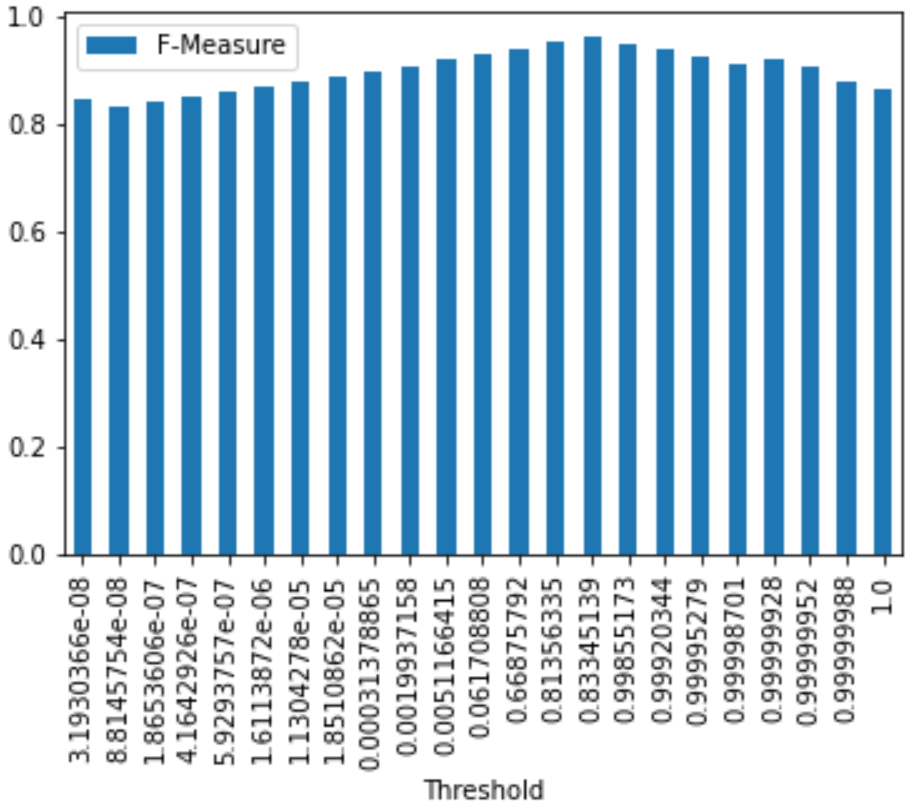}}
  \caption{Bar Graphs Determining Optimal Threshold - Highest values correspond to Optimal Threshold}
\end{figure}

Using the optimal threshold leads to the following conclusions:

\begin{itemize}[noitemsep]
  \item \textbf{False Positives} reduced from 4 to \textbf{2}.
  \item \textbf{Accuracy} improved from 92.0\% to \textbf{95.2\%}.
  \item \textbf{Specificity} improved from 81.8\% to \textbf{90.9\%}.
  \item \textbf{Precision} improved from 90.9\% to \textbf{95.2\%}.
  \item \textbf{Area under ROC} improved from 0.89 to \textbf{0.94}.
  \item \textbf{Cohen's Kappa} score improved from 0.81 to \textbf{0.89}.
  \item \textbf{F1 score} improved from 0.94 to \textbf{0.96}.
\end{itemize}

The important measurements using the optimal threshold are tabulated in Table 8, and so are the confusion matrices in Figure 12. A comparison of our results with similar works is depicted in Table 9.

\begin{table}[!ht]
\caption{Performance Results with Optimal Threshold}
\centering
\begin{tabular}{ c c | c c }
\hline
Category & Result & Metric & Result \\ [0.5ex] 
\hline
True Positives & 40 & Accuracy & 95.2\%\\
True Negatives & 20 & Specificity & 90.9\%\\
False Positives & 2 & Sensitivity & 97.5\%\\
False Negatives & 1 & Precision & 95.2\%\\
\hline
\end{tabular}
\end{table}

\begin{figure}[!ht]
  \centering
  \subfloat[Validation Set]{\includegraphics[scale = 0.34]{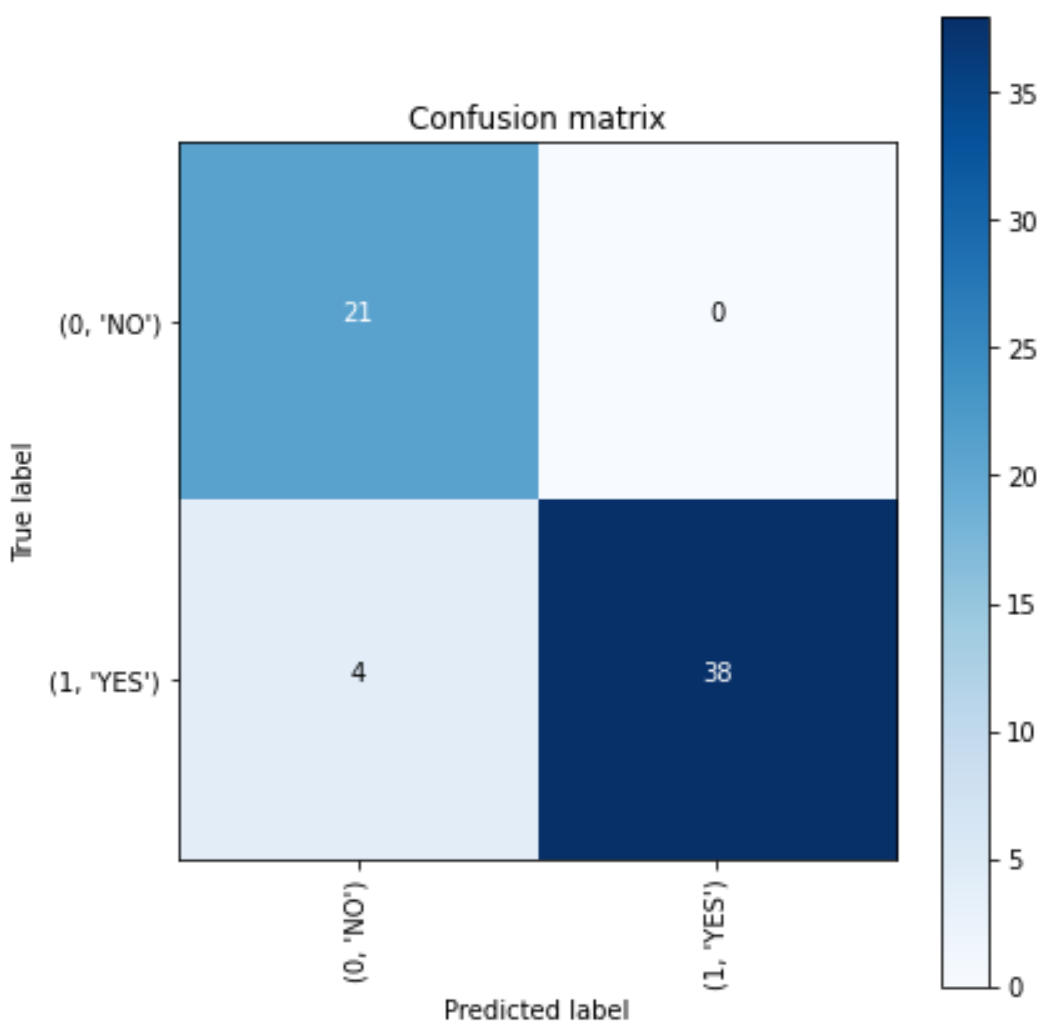}}
\hspace{2 cm}
  \subfloat[Test Set]{\includegraphics[scale = 0.34]{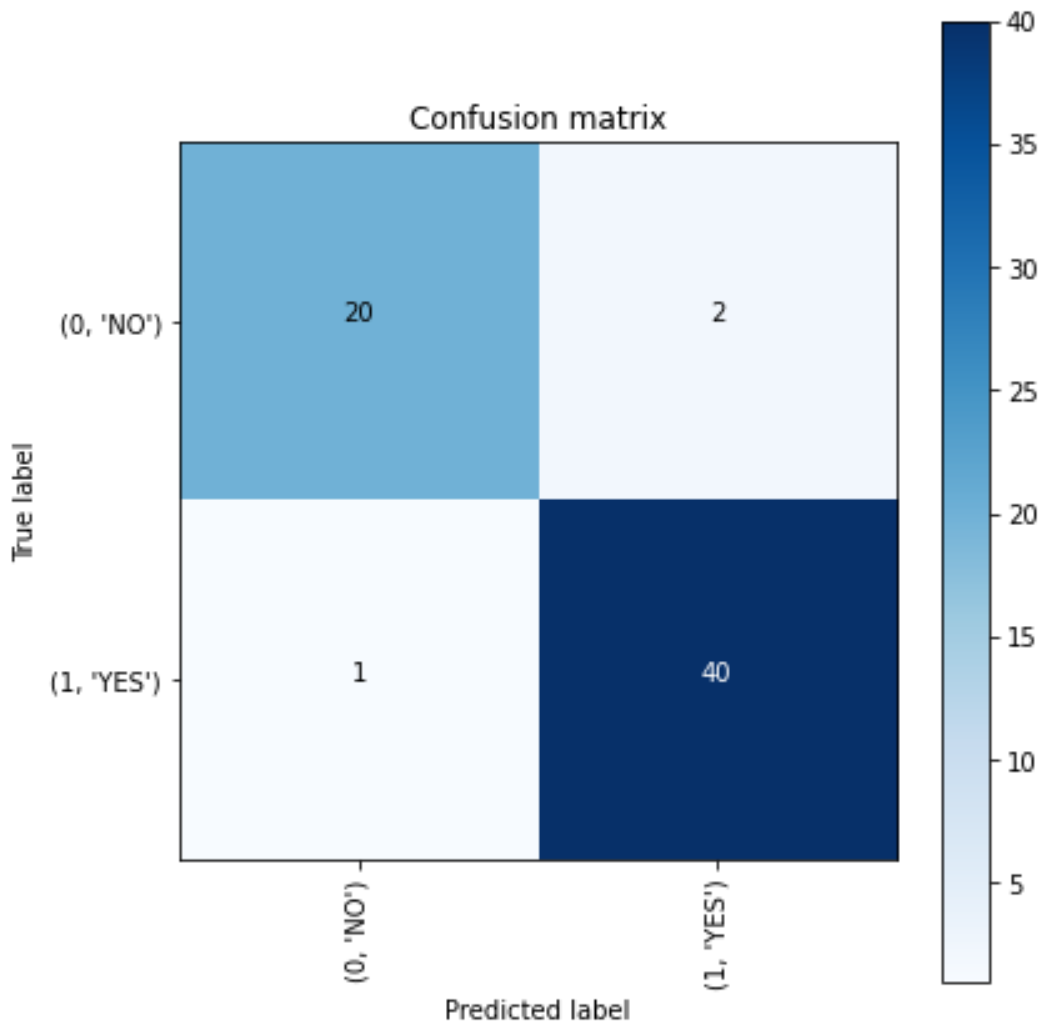}}
  \caption{Optimal Threshold Confusion Matrices}
\end{figure}

\begin{table}[!ht]
\caption{Comparison of proposed study with similar works}
\centering
\begin{tabular}{c p{20mm} c c c}
\hline
Study & Method & Accuracy (\%)& Sensitivity (\%) & Specificity (\%)\\  [0.5ex]
\hline
\textbf{Proposed Study} & \textbf{VGG16 with Transfer Learning} & \textbf{95.2} & \textbf{97.5} & \textbf{90.9}\\
\hline
Prashanth et al. \cite{LS3} & SVM with Striatal Binding Ratio values & 96.14 & 95.74 & 77.35\\
\hline
Brahim et al. \cite{brahim2017proposed} & PCA and SVM & 92.6 & 91.2 & 93.1\\
\hline
Rumman et al. \cite{rumman2018early} & Custom ANN & 94 & 100 & 88\\
\hline
Quan et al. \cite{quan2019datscan} & InceptionV3 with Transfer Learning & 98.4 & 98.8 & 97.6\\
\hline
Ortiz et al. \cite{ortiz2019parkinson} & LeNet-based & 95$\pm$0.3 & 94$\pm$0.4 & 95$\pm$0.4\\
\hline
Ortiz et al. \cite{ortiz2019parkinson} & AlexNet-based & 95$\pm$0.3 & 95$\pm$0.5 & 95$\pm$0.4\\
\hline
\end{tabular}
\end{table}

\section{Explainability of the Proposed Model using LIME}

\subsection{\textbf{Need for Interpretability}}

Artificial Intelligence solutions in the health care industry are mainly faced with the problem of \emph{explainability}. Questions such as \emph{"Why should I trust the outcome of this prediction?"} or \emph{"How did this program arrive at this diagnostic conclusion?"} need to be answered for medical workers to completely embrace the use of machine learning techniques in assisting them with early diagnosis. No matter how accurate a model is, it needs to be able to produce an argument explaining why the algorithm came up with a certain prediction or suggestion. While some models like decision trees are transparent, the current state-of-the-art models in the vast majority of AI applications in healthcare are neural networks that are \emph{black box} in nature and lack any kind of explanations for their predictions. This poses as a potential risk in situations where the stakes are high, such as a patient's re-admission to a hospital or determining the end of life care support for a patient. Recent efforts to develop the explainability of these black box models come under the research area of \emph{Explainable AI} and include works such as DeepLIFT \cite{shrikumar2017learning}, RISE \cite{petsiuk2018rise}, SHAP \cite{lundberg2017unified} and finally LIME \cite{ribeiro2016should}, which this study uses to interpret its results. \par

\begin{figure}[!ht]
\includegraphics[trim=0 100 0 100, clip, width=\textwidth]{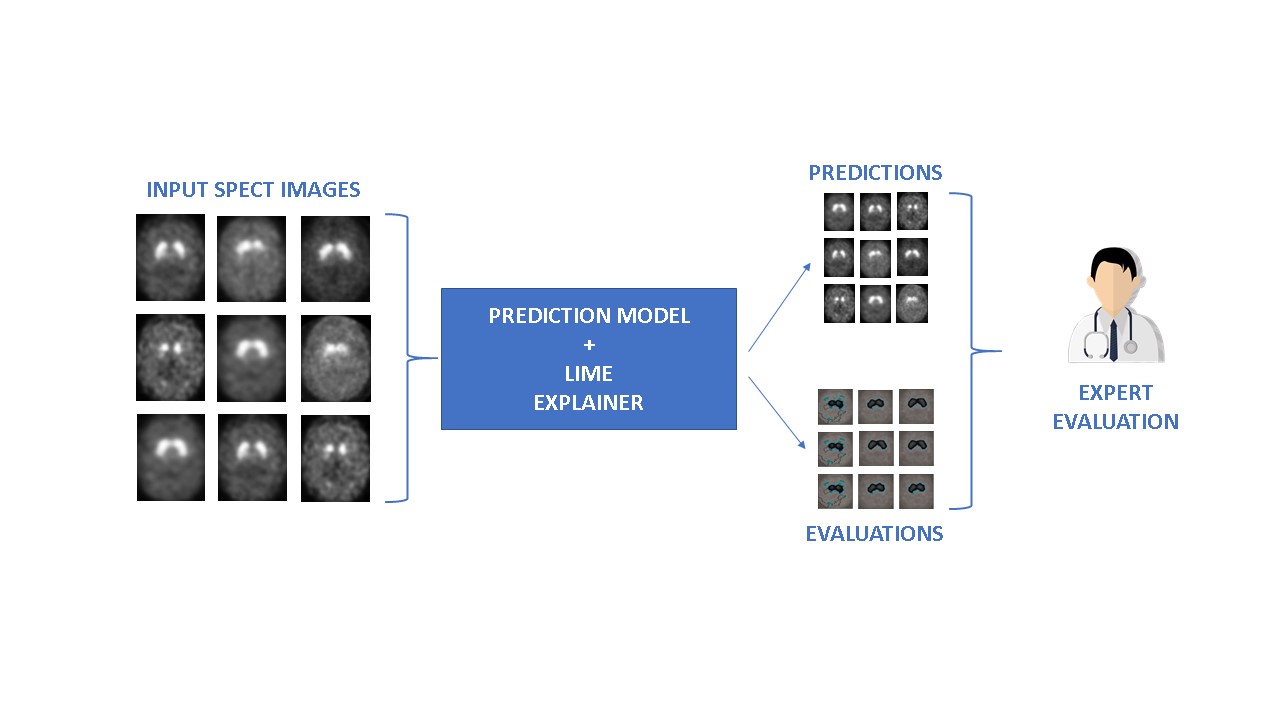}
\caption{An illustration of patient diagnosis by an expert using model predictions and their corresponding LIME explanations}
\end{figure}

In a previous experiment \cite{lapuschkin2016analyzing} where \emph{Fisher Vector} classifiers were used for the task of image recognition \cite{sanchez2013image} and an interpretability technique called 'Layer-wise Relevance Propagation' (LRP) \cite{bach2015pixel} was applied to decrypt the model's predictions, a peculiar observation was made. It was found that in specific cases where the input image consisted of a \emph{"horse"} the model was weighing its decision primarily not on any of the horse's physical features, but on a certain \emph{copyright tag} present on the bottom left of the image which turned out to be a characteristic of all the horse images used in training. This was certainly an egregious error on the part of the model and such an example certainly depicts the need for interpretability of deep learning models especially in the medical field where such mistakes cannot be allowed to happen. 

\subsection{\textbf{Local Interpretable Model-Agnostic Explanations (LIME)}}

The LIME framework is essentially a \emph{local surrogate model} which is an interpretable framework, utilised to explain independent predictions of \emph{'black box'} (i.e. underlying working is hidden) machine learning models \cite{molnar2019interpretable}. LIME conducts tests on what would happen to the predictions of the model when the user provides alterations of their data into the model. LIME, in this principle, engenders a novel dataset comprising of permuted specimens and the analogous predictions of the black box model. On this novel dataset, the framework then trains an interpretable model (e.g linear regression model, decision tree, etc.), that is weighted by the closeness of the sampled instances, to the instance of concern which is required to be explained. The learned model must be a plausible estimate of the machine learning model's predictions locally. Arithmetically, local surrogate models with the interpretability constraint can be depicted as follows in equation(4): 

\begin{equation}
interpretation(x) =  arg\ min_{v \in V}\ L(u,v,\pi_x) + \omega (v)
\end{equation}
\\
We consider an explainable model \emph{v} (e.g. decision tree) for the sample \emph{x} which will reduce a loss \emph{L} (e.g. binary cross entropy), and meters how near the interpretation is, relative to the predicted value of the initial model \emph{u} (e.g. a neural network model). This process is done all while keeping the model intricacy $\omega(g)$ minimum. Here, \emph{V} is the collection of realizable explanations, for which in a hypothetical case, may be feasible decision tree models. The closeness measure $\pi_x$ defines the extent of the locality around sample \emph{x}, and is what we consider for the explanation. \par

\begin{figure}[!ht]
\includegraphics[trim=0 200 0 200, clip, width=\textwidth]{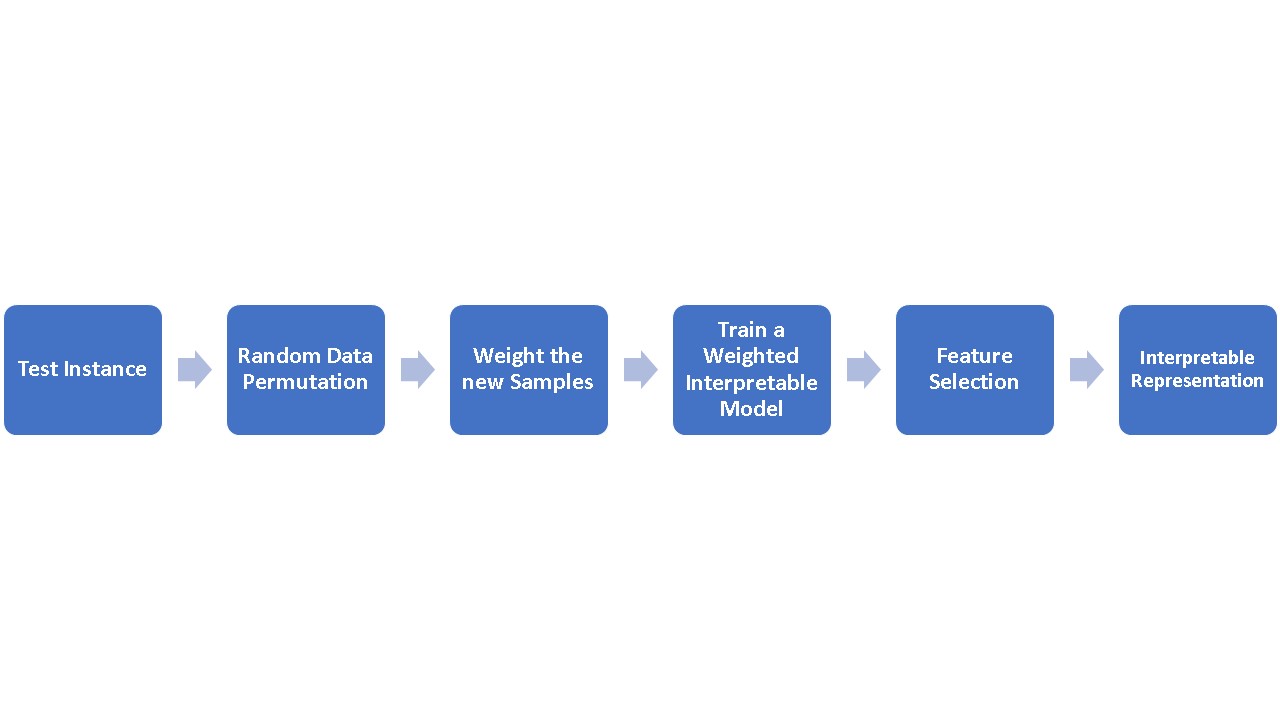}
\caption{Block diagram of how LIME works}
\end{figure}

For explaining images, permutations of the images are developed by segregating the image into \emph{superpixels} and switching these superpixels on or off. Superpixels are interlinked pixels with analogous colors and can be switched off by replacing every pixel with a user-specified one. The user may also specify a numerical likelihood for switching off a superpixel in each variation sample so that they may observe only the highest contributing factors. The use of these superpixels for explaining the decisions on the DaTscan images are discussed in the next section.

\subsection{\textbf{Interpretation of DaTscans}}

The region of interest (ROI) in our data are the \emph{putamen} and \emph{caudate} regions of the brain and hence the LIME explainer instance attributes the superpixels relating to these regions as the portions of the image with the highest weights or influence in determining the outcome of the prediction. The application of LIME as seen on the samples in Figure 15 and Figure 16 allows the visual tracing of the ROI which makes it easier for non-experts in the field to determine the diagnosis of the patient. \par

\begin{figure}[!ht]
  \subfloat{
	\begin{minipage}[c][1\width]{
	   0.145\textwidth}
	   \centering
	   \includegraphics[width=1\textwidth, height=2.4cm]{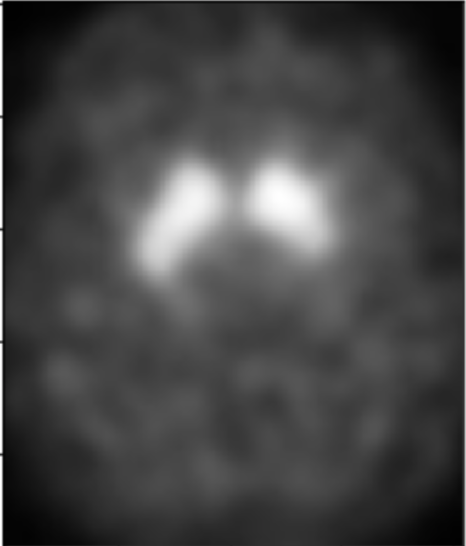}
	\end{minipage}}
 \hfill 	
  \subfloat{
	\begin{minipage}[c][1\width]{
	   0.145\textwidth}
	   \centering
	   \includegraphics[width=1\textwidth, height=2.4cm]{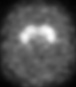}
	\end{minipage}}
 \hfill	
  \subfloat{
	\begin{minipage}[c][1\width]{
	   0.145\textwidth}
	   \centering
	   \includegraphics[width=1\textwidth, height=2.4cm]{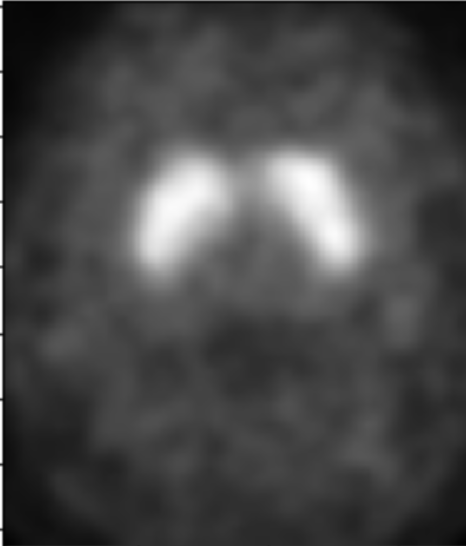}
	\end{minipage}}
\hfill	
  \subfloat{
	\begin{minipage}[c][1\width]{
	   0.145\textwidth}
	   \centering
	   \includegraphics[width=1\textwidth, height=2.4cm]{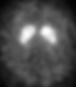}
	\end{minipage}}
\hfill	
  \subfloat{
	\begin{minipage}[c][\width]{
	   0.145\textwidth}
	   \centering
	   \includegraphics[width=\textwidth, height=2.4cm]{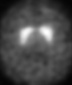}
	\end{minipage}}
\hfill	
  \subfloat{
	\begin{minipage}[c][\width]{
	   0.145\textwidth}
	   \centering
	   \includegraphics[width=1\textwidth, height=2.4cm]{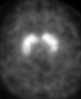}
	\end{minipage}}
	
	  \subfloat{
	\begin{minipage}[c][1\width]{
	   0.145\textwidth}
	   \centering
	   \includegraphics[width=1\textwidth]{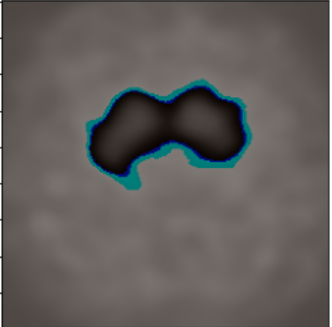}
	   \caption*{(a) Patient-1 (HC)}
	\end{minipage}}
 \hfill 	
  \subfloat{
	\begin{minipage}[c][1\width]{
	   0.145\textwidth}
	   \centering
	   \includegraphics[width=1\textwidth]{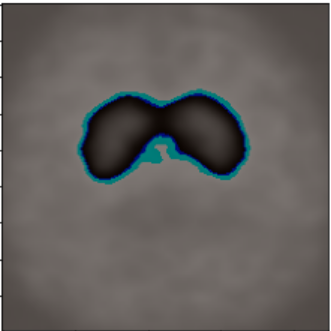}
	   \caption*{(b) Patient-2 (HC)}
	\end{minipage}}
 \hfill	
  \subfloat{
	\begin{minipage}[c][1\width]{
	   0.145\textwidth}
	   \centering
	   \includegraphics[width=1\textwidth]{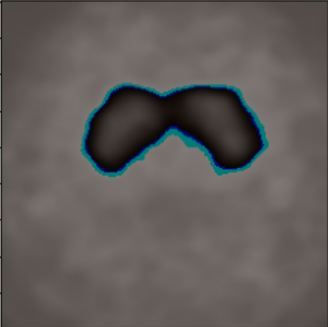}
	   \caption*{(c) Patient-3 (HC)}
	\end{minipage}}
\hfill	
  \subfloat{
	\begin{minipage}[c][1\width]{
	   0.145\textwidth}
	   \centering
	   \includegraphics[width=1\textwidth]{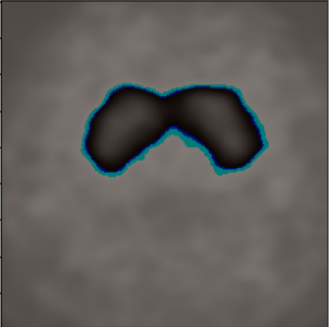}
	   \caption*{(d) Patient-4 (HC)}
	\end{minipage}}
\hfill	
  \subfloat{
	\begin{minipage}[c][\width]{
	   0.145\textwidth}
	   \centering
	   \includegraphics[width=\textwidth]{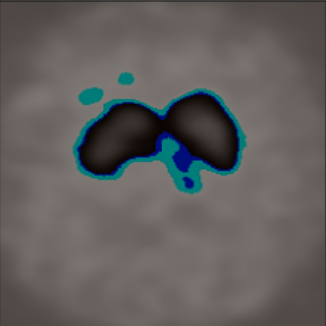}
	   \caption*{(e) Patient-5 (HC)}
	\end{minipage}}
\hfill	
  \subfloat{
	\begin{minipage}[c][\width]{
	   0.145\textwidth}
	   \centering
	   \includegraphics[width=1\textwidth]{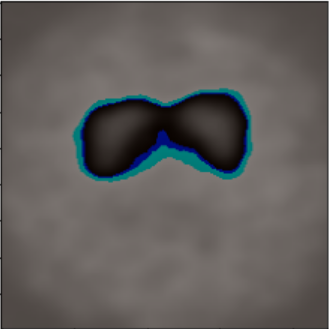}
	   \caption*{(f) Patient-6 (HC)}
	\end{minipage}}
	\bigskip
\caption{Samples of HC classified interpretations}
\end{figure}
\begin{figure}[!ht]
  \subfloat{
	\begin{minipage}[c][1\width]{
	   0.145\textwidth}
	   \centering
	   \includegraphics[width=1\textwidth, height=2.4cm]{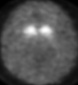}
	  
	\end{minipage}}
 \hfill 	
  \subfloat{
	\begin{minipage}[c][1\width]{
	   0.145\textwidth}
	   \centering
	   \includegraphics[width=1\textwidth, height=2.4cm]{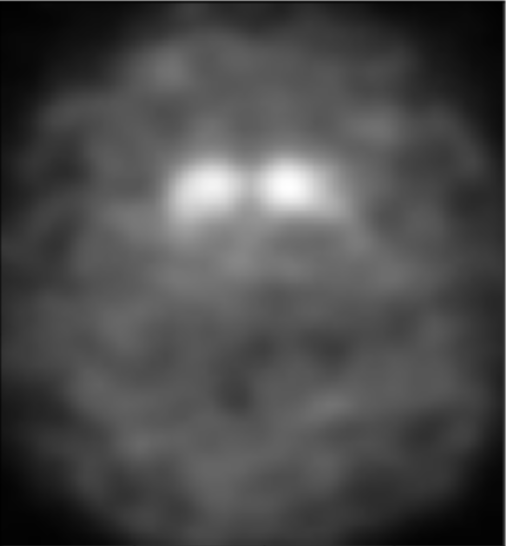}
	   
	\end{minipage}}
 \hfill	
  \subfloat{
	\begin{minipage}[c][1\width]{
	   0.145\textwidth}
	   \centering
	   \includegraphics[width=1\textwidth, height=2.4cm]{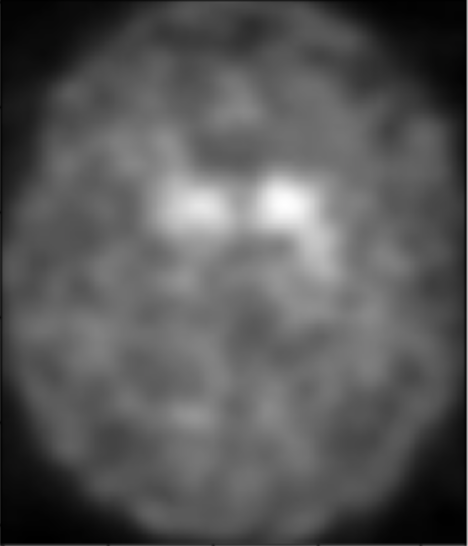}
	\end{minipage}}
\hfill	
  \subfloat{
	\begin{minipage}[c][1\width]{
	   0.145\textwidth}
	   \centering
	   \includegraphics[width=1\textwidth, height=2.4cm]{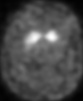}
	\end{minipage}}
\hfill	
  \subfloat{
	\begin{minipage}[c][\width]{
	   0.145\textwidth}
	   \centering
	   \includegraphics[width=\textwidth, height=2.4cm]{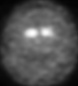}
	\end{minipage}}
\hfill	
  \subfloat{
	\begin{minipage}[c][\width]{
	   0.145\textwidth}
	   \centering
	   \includegraphics[width=1\textwidth, height=2.4cm]{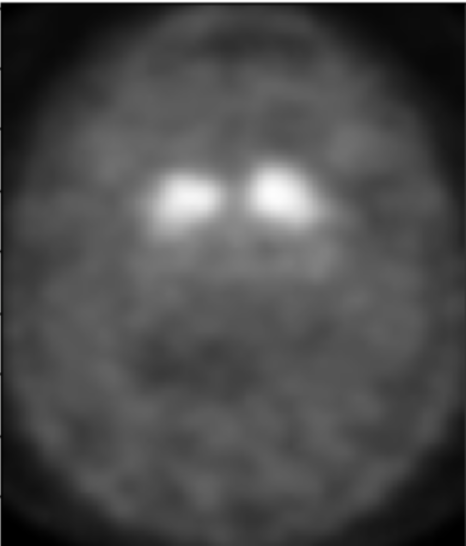}
	\end{minipage}}
	
	  \subfloat{
	\begin{minipage}[c][1\width]{
	   0.145\textwidth}
	   \centering
	   \includegraphics[width=1\textwidth]{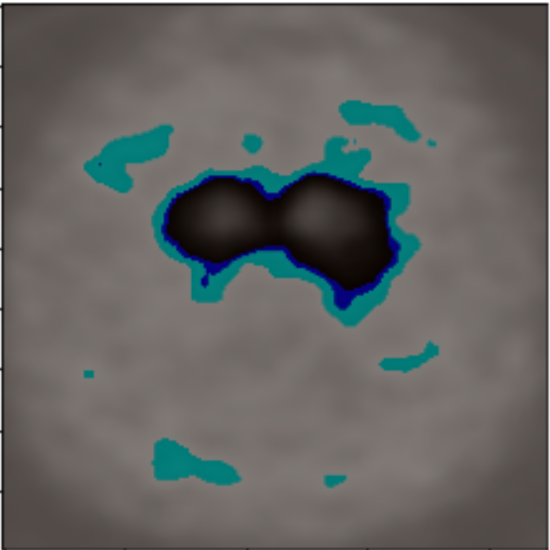}
	    \caption*{(a) Patient-1 (PD)}
	\end{minipage}}
 \hfill 	
  \subfloat{
	\begin{minipage}[c][1\width]{
	   0.145\textwidth}
	   \centering
	   \includegraphics[width=1\textwidth]{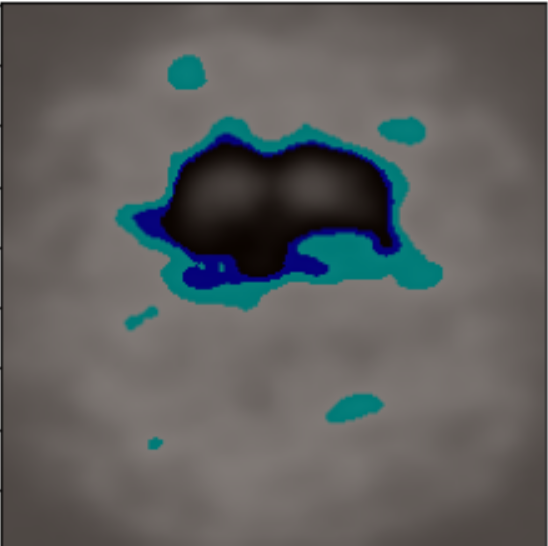}
	    \caption*{(b) Patient-2 (PD)}
	\end{minipage}}
 \hfill	
  \subfloat{
	\begin{minipage}[c][1\width]{
	   0.145\textwidth}
	   \centering
	   \includegraphics[width=1\textwidth]{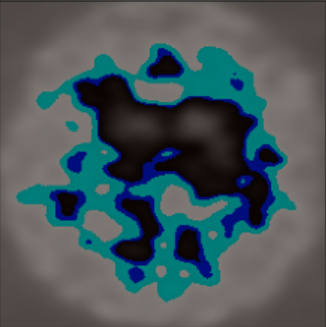}
	    \caption*{(c) Patient-3 (PD)}
	\end{minipage}}
\hfill	
  \subfloat{
	\begin{minipage}[c][1\width]{
	   0.145\textwidth}
	   \centering
	   \includegraphics[width=1\textwidth]{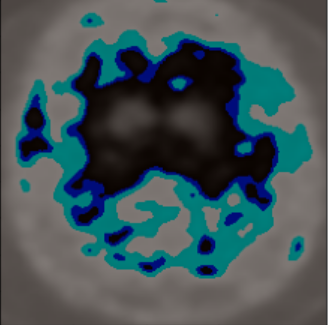}
	    \caption*{(d) Patient-4 (PD)}
	\end{minipage}}
\hfill	
  \subfloat{
	\begin{minipage}[c][\width]{
	   0.145\textwidth}
	   \centering
	   \includegraphics[width=\textwidth]{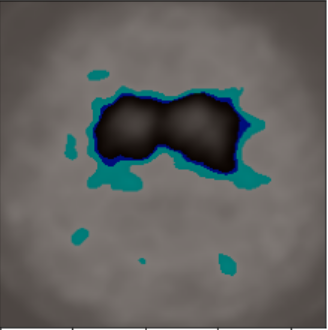}
	    \caption*{(e) Patient-5 (PD)}
	\end{minipage}}
\hfill	
  \subfloat{
	\begin{minipage}[c][\width]{
	   0.145\textwidth}
	   \centering
	   \includegraphics[width=1\textwidth]{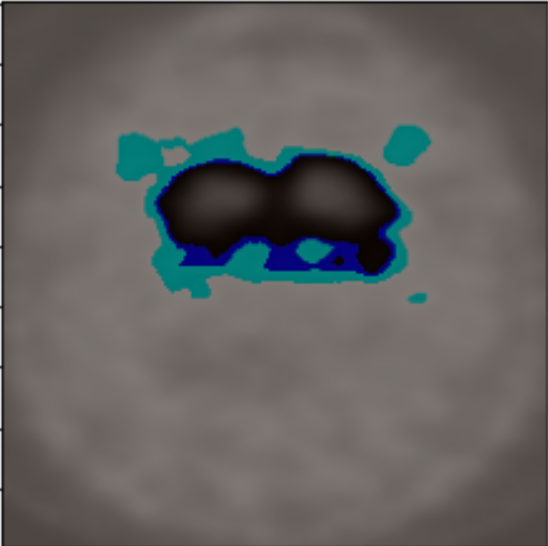}
	    \caption*{(f) Patient-6 (PD)}
	\end{minipage}}
	\bigskip
\caption{Samples PD classified interpretations}
\end{figure}

As seen in the two examples from Figure 17, the LIME explainer emphasised the \emph{healthy} putamen and caudate regions of a non-PD patient to be the influencing regions in classifying the data as healthy control. Figure 17(a) and 17(c) depict the SPECT scans (after preprocessing) and Figure 17(b) and 17(d) depict the corresponding output after applying LIME. \par

\begin{figure}[!ht]
  \centering
  \subfloat[Raw Image]{\includegraphics[scale = 0.3]{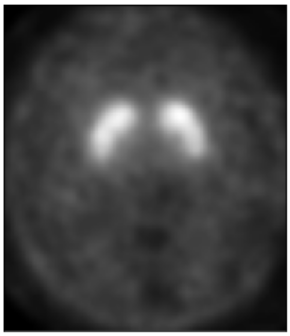}}
\hspace{1 cm}
  \subfloat[With LIME]{\includegraphics[scale = 0.3]{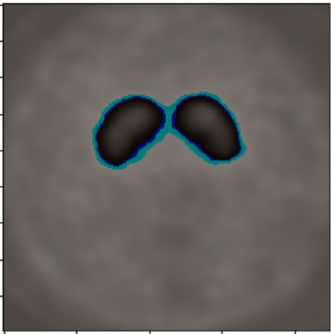}}
  \\
  \subfloat[Raw Image]{\includegraphics[scale = 0.3]{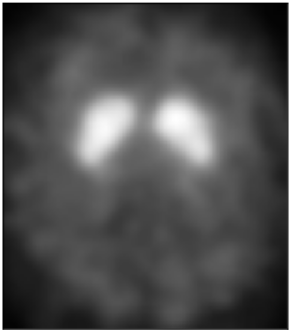}}
\hspace{1 cm}
  \subfloat[With LIME]{\includegraphics[scale = 0.3]{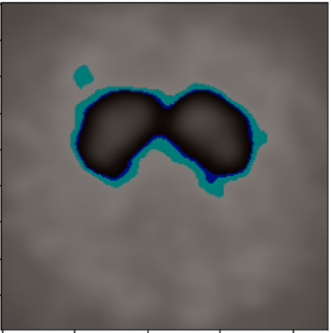}}
  \caption{Healthy Control}
\end{figure}

\begin{figure}[!ht]
  \centering
  \subfloat[Raw Image]{\includegraphics[scale = 0.3]{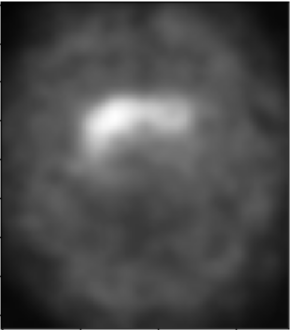}}
\hspace{1 cm}
  \subfloat[With LIME]{\includegraphics[scale = 0.3]{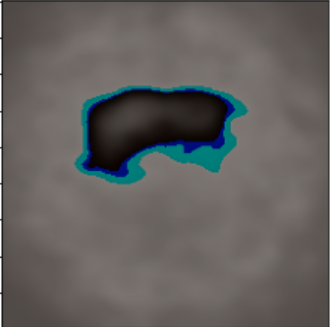}}
  \\
  \subfloat[Raw Image]{\includegraphics[scale = 0.3]{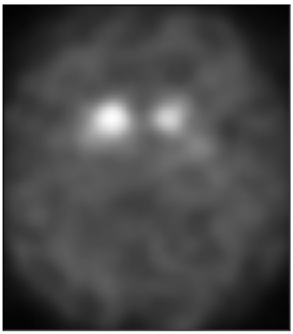}}
\hspace{1 cm}
  \subfloat[With LIME]{\includegraphics[scale = 0.3]{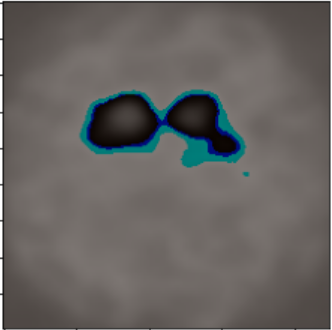}}
  \caption{Parkinson's Disease}
\end{figure}

Similarly seen in the two examples from Figure 18, the LIME explainer emphasised the \emph{abnormal} or \emph{reduced} features of the  putamen and caudate regions of a non-PD patient to be the influencing regions in classifying the data as PD. Figure 18(a) and 18(c) depict the SPECT scans (after preprocessing) and  Figure 18(b) and 18(d) depict the corresponding output after applying LIME. We may see that in the explanations of a few samples like (c) and (d) of Figure 16, the emphasized superpixels are more distorted. This explanation could be the result of an anomalous increased dopamine activity in nearby areas of the ROI, a characteristic feature of late-stage PD. Smaller ROIs have most probably prompted the model to learn PD relevant features surrounding the putamen and caudate regions as well and hence we observe non-uniform superpixel distribution among such PD classification explanations.

\section{Conclusion}

The purpose of this study was to classify SPECT DaTscan images as having Parkinson's disease or not while providing meaningful insights into the decisions generated by the model. Using the VGG16 CNN architecture along with transfer learning, the model was able to achieve an accuracy of 95.2\%. This study aimed at making an early diagnosis for Parkinson's disease faster, more intuitive, and is proposed to be applied in real-world scenarios. \par

These results lay the groundwork for future studies where a larger dataset may be available and the extent of the class imbalance may be smaller. Model accuracy has scope for improvement, thereby reducing the number of false positives and negatives, through possible tuning of hyper-parameters or using different network architectures. Improvements in neural network input limitations may allow an entire 3D volume image (as is the case with most brain scans) to be trained on a model, and not just a single slice of the volume, hence preserving relationships amongst any possibly important adjacent slices. Another possible shortcoming is the authenticity of the labelling of the obtained data before training the model. Lack of a definitive diagnostic test for Parkinson's disease means that the data labelling is still of questionable accuracy being subjective to the assessment of human evaluations. The model may also need to undergo clinical validation and tested in real-time with novel data. \par

Finally, we can conclude that a model with reliable accuracy was developed on a sample size that was sufficiently large and diverse. It utilises an effective approach, saving valuable time and resources for healthcare workers. The study assists in the early diagnosis of Parkinson's disease through explanations, thereby developing confidence in the use of computer-aided diagnosis for medical purposes.

\bibliography{image_draft}

\end{document}